\documentclass[sigconf]{acmart}
\AtBeginDocument{%
  }

\copyrightyear{2023}
\acmYear{2023}
\setcopyright{acmlicensed}
\acmConference[MM '23] {Proceedings of the 31st ACM International Conference on Multimedia}{October 29--November 3, 2023}{Ottawa, ON, Canada.}
\acmBooktitle{Proceedings of the 31st ACM International Conference on Multimedia (MM '23), October 29--November 3, 2023, Ottawa, ON, Canada}
\acmPrice{15.00}
\acmDOI{10.1145/3581783.3611752}
\acmISBN{979-8-4007-0108-5/23/10}

\acmSubmissionID{266}

\usepackage{graphicx}
\usepackage{amsmath}
\usepackage{booktabs}
\usepackage{bbding}
\usepackage{subcaption} 
\usepackage{multirow} 
\usepackage{balance}

\usepackage[ruled,vlined]{algorithm2e}

\begin{document}

\title{Uncertainty-Guided Spatial Pruning Architecture for Efficient Frame Interpolation}



\author{Ri Cheng}
\orcid{0000-0002-5866-6847}
\affiliation{%
  \institution{School of Computer Science, Shanghai Key Laboratory of Intelligent Information Processing, Shanghai Collaborative Innovation Center of Intelligent Visual Computing, Fudan University}
  \city{Shanghai}
  \country{China}
}
\email{rcheng22@m.fudan.edu.cn}

\author{Xuhao Jiang}
\orcid{0000-0002-4646-5052}
\affiliation{%
  \institution{School of Computer Science, Shanghai Key Laboratory of Intelligent Information Processing, Shanghai Collaborative Innovation Center of Intelligent Visual Computing, Fudan University}
  \city{Shanghai}
  \country{China}
}
\email{20110240011@fudan.edu.cn}

\author{Ruian He}
\orcid{0000-0001-9598-3043}
\affiliation{%
  \institution{School of Computer Science, Shanghai Key Laboratory of Intelligent Information Processing, Shanghai Collaborative Innovation Center of Intelligent Visual Computing, Fudan University}
  \city{Shanghai}
  \country{China}
}
\email{rahe16@fudan.edu.cn}

\author{Shili Zhou}
\orcid{0000-0001-7283-2314}
\affiliation{%
  \institution{School of Computer Science, Shanghai Key Laboratory of Intelligent Information Processing, Shanghai Collaborative Innovation Center of Intelligent Visual Computing, Fudan University}
  \city{Shanghai}
  \country{China}
}
\email{slzhou19@fudan.edu.cn}

\author{Weimin Tan}
\orcid{0000-0001-7677-4772}
\authornote{Corresponding Author. This work is supported by NSFC (Grant No.: U2001209) and Natural Science Foundation of Shanghai (21ZR1406600).}
\affiliation{%
  \institution{School of Computer Science, Shanghai Key Laboratory of Intelligent Information Processing, Shanghai Collaborative Innovation Center of Intelligent Visual Computing, Fudan University}
  \city{Shanghai}
  \country{China}
}
\email{wmtan@fudan.edu.cn}

\author{Bo Yan}
\orcid{0000-0003-0256-9682}
\authornotemark[1]
\affiliation{%
  \institution{School of Computer Science, Shanghai Key Laboratory of Intelligent Information Processing, Shanghai Collaborative Innovation Center of Intelligent Visual Computing, Fudan University}
  \city{Shanghai}
  \country{China}
}
\email{byan@fudan.edu.cn}

\renewcommand{\shortauthors}{Ri Cheng et al.}

\begin{abstract}
The video frame interpolation (VFI) model applies the convolution operation to all locations, leading to redundant computations in regions with easy motion. 
We can use dynamic spatial pruning method to skip redundant computation, but this method cannot properly identify easy regions in VFI tasks without supervision. In this paper, we develop an Uncertainty-Guided Spatial Pruning (UGSP) architecture to skip redundant computation for efficient frame interpolation dynamically. Specifically, pixels with low uncertainty indicate easy regions, where the calculation can be reduced without bringing undesirable visual results. Therefore, we utilize uncertainty-generated mask labels to guide our UGSP in properly locating the easy region. Furthermore, we propose a self-contrast training strategy that leverages an auxiliary non-pruning branch to improve the performance of our UGSP. Extensive experiments show that UGSP maintains performance but reduces FLOPs by 34\%/52\%/30\% compared to baseline without pruning on Vimeo90K/UCF101/MiddleBury datasets. In addition, our method achieves state-of-the-art performance with lower FLOPs on multiple benchmarks.
\end{abstract}

\ccsdesc[500]{Computing methodologies~Reconstruction}

\keywords{Video Frame Interpolation, Dynamic Network, Spatial Pruning, Uncertainty}

\maketitle

\section{Introduction}
\label{sec:intro}

    \begin{figure}[t]
    \centering
    \includegraphics[width=\linewidth]{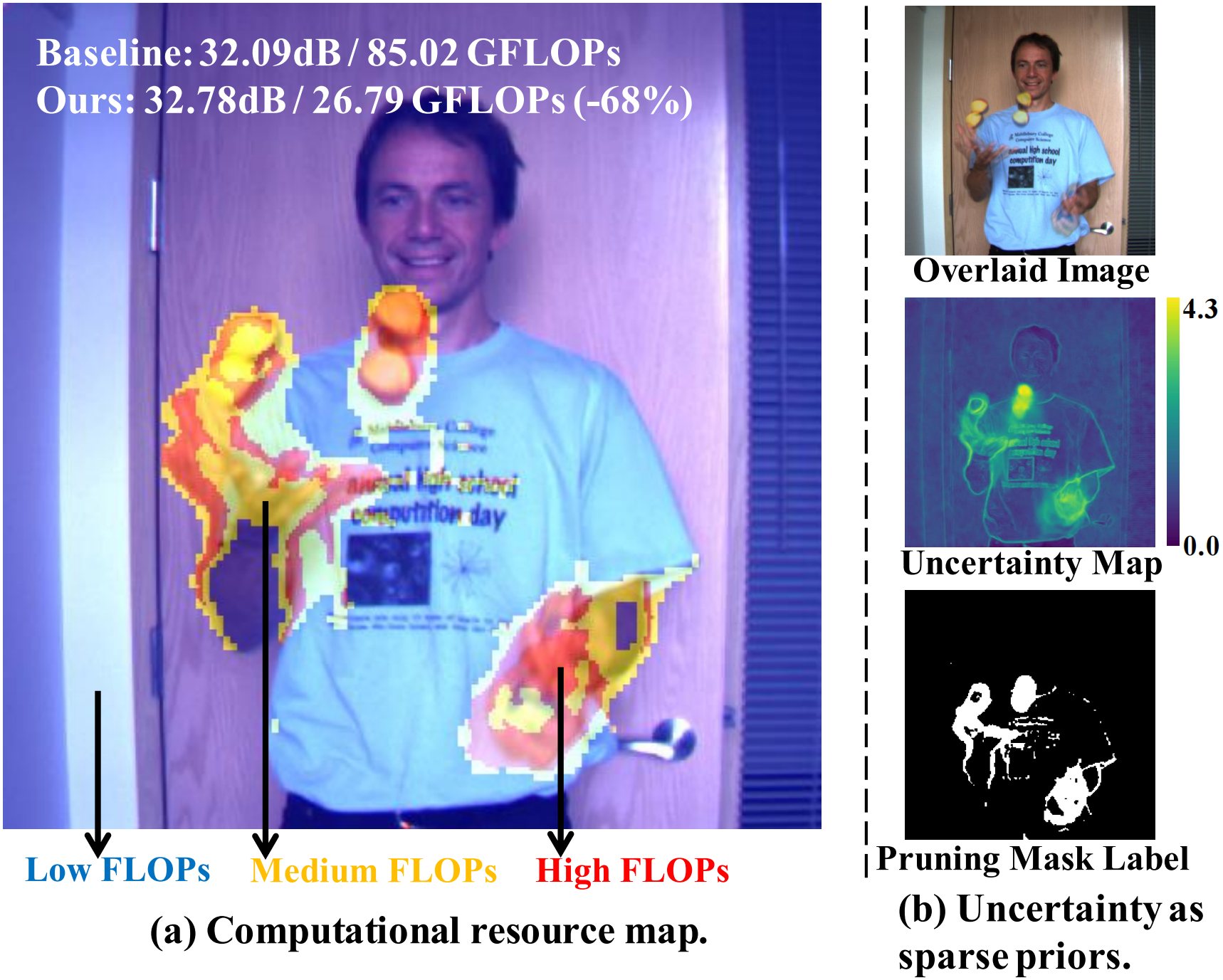}
    \caption{   (a) 94\% of regions require low computational costs. By using our method for baseline pruning, we maintain performance while reducing FLOPs by 68\%.
    (b) From top to bottom: overlaid image, the uncertainty map generated by our model, and the uncertainty-generated pruning mask label.
    }
    \label{001_intro}
    \end{figure}

Video frame interpolation (VFI) attempts to interpolate intermediate frames in the middle of two input consecutive frames.
Applications such as slow-motion generation \cite{Jiang_2018_CVPR}, video compression \cite{Wu_2018_ECCV}, and novel view synthesis \cite{10.1007/978-3-319-46493-0_18}, utilize frame interpolation frameworks extensively.
In recent years, well-designed deep learning-based model shows excellent progress in VFI task due to its powerful feature representation ability.
When performing such deep VFI models on the resource-limited edge devices, it encounters a computational resource constraint issue.
Consequently, efficient VFI is a realistic requirement for these devices.

Real-world videos contain large and non-linear
motion that requires a large number of convolution layers to expand the receptive field, resulting in significant computational demand.
To address this problem, several efforts have been made by utilizing coarse-to-fine structure \cite{huang2022rife, Park_2021_ICCV,Lee_2020_CVPR} or predicting the intermediate flow and frame in one step \cite{Kong_2022_CVPR}.
However, these networks still involve redundant computation since they apply the convolution operation to all locations equally without distinguishing between challenging and easy regions.
Based on the observation in Section~\ref{sec:ob} that for easy regions with static or straightforward motion,  convolutions in coarse scales are sufficient to achieve satisfactory outcomes.
What is more, these easy regions occupy a large portion of the input frames.
Therefore, by reducing the heavy computation in these regions, model inference speed can be significantly increased. As shown in Figure~\ref{001_intro}(a), by reducing the redundant calculations in static regions (colored blue), we can save 68\% FLOPs. In addition, our pruning model has a higher PSNR since the baseline model cannot estimate the trajectory of the ball accurately, indicating that our model can prioritize the reconstructed quality of challenging areas by using the pruning mask.


In this paper, we propose an Uncertainty-Guided Spatial Pruning (UGSP) architecture to dynamically skip redundant computations for efficient frame interpolation.
During inference, UGSP predicts pruning masks to localize redundant computation regions and skip them using sparse convolution \cite{Wang_2021_CVPR}.
However, we discover that directly predicting such a mask without supervision is inaccurate and unstable.
Aleatoric uncertainty \cite{NIPS2017_2650d608,NEURIPS2021_88a19961} measures the prediction difficulty of each pixel in an instance, so it can be used to guide the estimation of easy regions that require less computation resources.
Therefore, we propose using uncertainty to enhance the performance of our VFI pruning model. 


Specifically, the UGSP training procedure consists of two phases.
In the first phase, the model is trained to predict the mean and variance of each pixel in the intermediate target frame.
We observe that pixels with large uncertainty (variation) denote complex and large movements, so they demand more computational resources, as shown at the middle of Figure~\ref{001_intro}(b). As observed in Section~\ref{sec:ob}, we find that large uncertainty regions are crucial to the final visual quality, since the improvement in visual quality will increase significantly when sufficient computational resources are allocated to these regions.
In this case, we propose an uncertainty-guided mask prediction approach
to guide the prediction of pruning mask using the uncertainty-generated mask label. An example of the mask label is shown at the bottom of Figure~\ref{001_intro}(b).

In the second phase, we supervise the estimation of the pruning mask using the mask label.
In addition, we design a self-contrast training strategy for enhancing the performance of VFI model estimating ability.
Specifically, an auxiliary non-pruning branch in UGSP generates the features of intermediate frame to guide training through our self-contrast loss. 
In summary, our main contributions are summarized as follows:


\begin{itemize}
\item We propose the Uncertainty-Guided Spatial Pruning (UGSP) architecture for dynamically accelerating VFI by reducing redundant computation. To our knowledge, we are the first to incorporate uncertainty into spatial pruning networks with promising results.

\item 
We observe low uncertainty occurs in easy movement regions where redundant computations exists, so we propose using uncertainty-generated mask labels to guide our framework in properly locating easy regions.
In addition, we propose a self-contrast training strategy that guides UGSP training using the interpolated frame features generated by the auxiliary non-pruning branch.

\item 
Our UGSP can reduce FLOPs by 34\%/52\%/30\% while maintaining PSNR performance to the baseline without pruning on Vimeo90K\cite{vimeo}/UCF101\cite{ucf101}/MiddleBury\cite{Baker2011} datasets.
The experimental results show that UGSP achieves the best performance with lower FLOPs compared to state-of-the-arts.

\end{itemize}

\section{Related Work}
\label{sec:related_work}

\noindent
\textbf{Efficient Video Frame Interpolation.}
VFI aims to estimate the motion between two input frames and then interpolate one or more intermediate frames. Recent efforts on VFI primarily utilizes kernel-based \cite{Niklaus_2017_CVPR,Niklaus_2017_ICCV,Cheng_Chen_2020,Lee_2020_CVPR,Reda_2018_ECCV,9501506,Peleg_2019_CVPR,Bao_2019_CVPR,8840983}, flow-based \cite{Niklaus_2018_CVPR,Niklaus_2020_CVPR,Park_2021_ICCV,Jiang_2018_CVPR,NEURIPS2019_d045c59a,10.1007/978-3-030-58583-9_7,Kim_Oh_Kim_2020}, and transformer-based \cite{Lu_2022_CVPR,Shi_2022_CVPR} approaches.
However, these approaches need too much redundant computation and inference delay to be viable for resource-limited devices.
In recent years, several works have proposed computationally efficient methods for addressing the issue of limited computing resources.
IFRNet \cite{Kong_2022_CVPR} builds an efficient encoder-decoder based network that requires no extra synthesis or refinement modules.
RIFE \cite{huang2022rife} decreases the inference time by directly estimating the intermediate flows with much better speed.
CDFI \cite{Ding_2021_CVPR} uses a static pruning method to reduce the same amount of computation at all locations, and MADA \cite{Choi_2021_ICCV} determines computational resources for each sub-image.

\begin{figure*}[t]
\centering
\includegraphics[width=\linewidth]{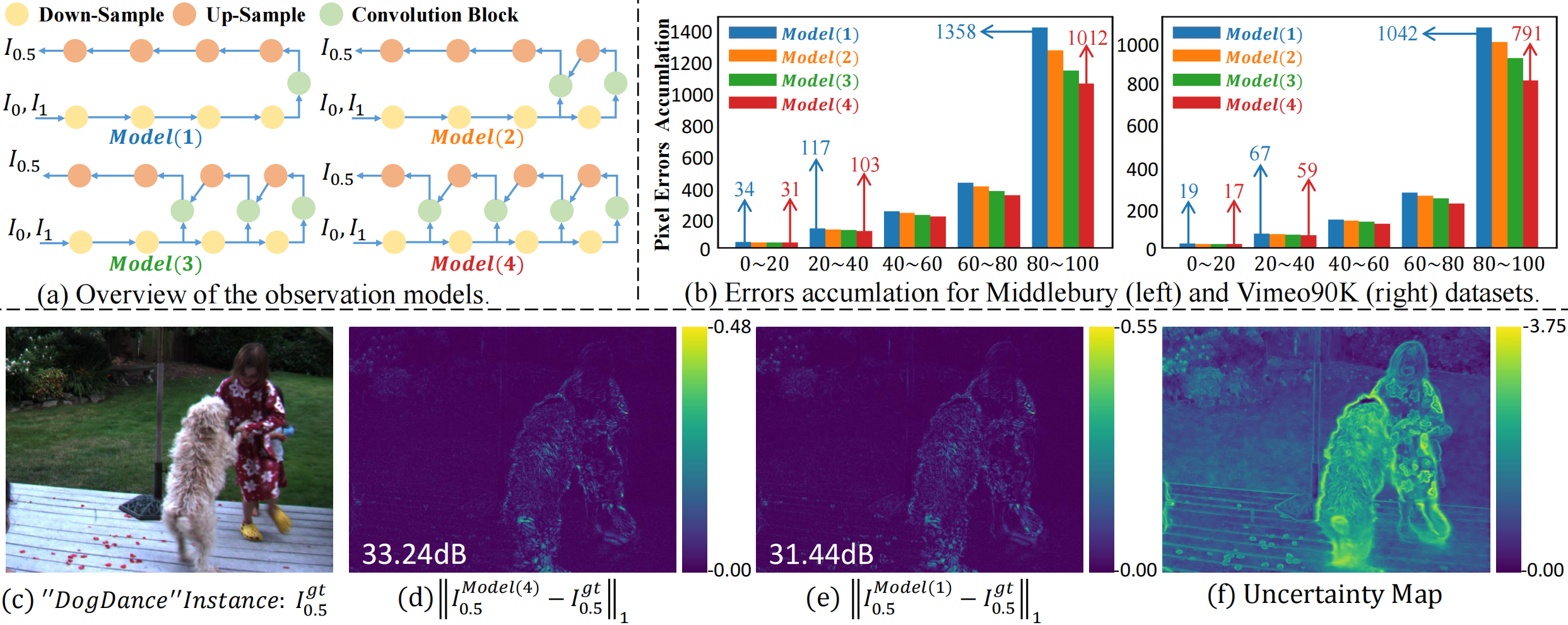}
\caption{ (a): Model(1), Model(2), Model(3), and Model(4) each contain 1, 2, 3 and 4 convolution blocks. (b): Y-axis: the accumulate of pixel errors between each Model and ground truth. X-axis: we rank pixel errors from small to large and divide them into five equal intervals based on the ranking.
(c) `DogDance' instance from Middlebury dataset.
(d-e): Display of the difference between Model(4) and ground truth, Model(1) and ground truth. (f) Uncertainty estimated by our uncertainty estimation network.
}
\label{002_ob}
\end{figure*}

\noindent
\textbf{Dynamic inference.}
Dynamic inference techniques \cite{9560049} can adapt the network structures during inference based on the input and consequently have advantageous properties such as efficiency, representation power, and interpretability.
Inference path selection \cite{Kong_2021_CVPR,Ding_2021_CVPR,Liu_2022_CVPR}, early stopping strategies \cite{10.1007/978-3-030-58517-4_17,pmlr-v70-bolukbasi17a,huang2018multi}, and adaptive skipping of redundant computation via sparse convolution \cite{Wang_2021_CVPR,Yang_2022_CVPR,10.1007/978-3-030-58452-8_31,Habibian_2021_CVPR,Parger_2022_CVPR} or explicitly skipping convolutions \cite{Wu_2018_CVPR,Mullapudi_2018_CVPR,2022arXiv220304845C} are related methods.
ClassSR \cite{Kong_2021_CVPR}, for instance, uses a classification network to identify the model channel number for each sub-image, while MADA \cite{Choi_2021_ICCV} identifies the model layer number and input scale for each sub-image.
However, due to the low resolution of the sub-image, the receptive field of convolution is restricted, and computing resources are roughly decreased at the patch level.
SMSR \cite{Wang_2021_CVPR} and QueryDet \cite{Yang_2022_CVPR} allocate computation resources for the important location using predicted masks and sparse convolution for the super-resolution and object detection tasks, respectively. However, their effectiveness is not demonstrated for VFI, and estimated masks are inaccurate as they are not supervised.




\noindent
\textbf{Uncertainty in Deep Learning.}
There are two primary types of uncertainty in deep learning models \cite{NIPS2017_2650d608}. Aleatoric uncertainty captures the noise inherent in observation data, and epistemic uncertainty accounts for the uncertainty of the model about its predictions. Uncertainty is widely used to increase the performance of deep-learning tasks such as face recognition \cite{Chang_2020_CVPR}, image classification \cite{NIPS2017_2650d608}, image segmentation \cite{7803544}, image denoising \cite{10096540}, video object segmentation \cite{Xu_Wang_Li_Lu_2022}, and super-resolution \cite{NEURIPS2021_88a19961,tmp}.
In the image super-resolution task, pixels with large certainty, such as texture and edge pixels, will be prioritized based on their importance to visual quality using uncertainty-driven loss \cite{NEURIPS2021_88a19961}.
However, uncertainty is rarely incorporated into dynamic networks, even though it can provide sparse priors. For example, low-uncertainty regions require lower computational demands, so convolutions can be skipped.

\section{Proposed Method}
\label{sec:proposed_framework}



\subsection{Observation}
\label{sec:ob}
We first illustrate our statistical observations regarding the VFI results of the Middlebury \cite{Baker2011} and Vimeo90K \cite{vimeo} training datasets.
These observations reveal the inherent sparsity of the VFI task and motivate us to design more effective VFI frameworks.
As illustrated in Figure~\ref{002_ob}(a), we use four observation models Model(1)$\sim$Model(4) that are created by eliminating 0$\sim$3 convolution blocks from the IFRnet \cite{Kong_2022_CVPR}. The convolution block accounts for the majority of the computational burden at each scale. Consequently, Model(1) has the least computing resources for each pixel location, followed by Model(2), Model(3), and Model(4).

We first calculate the pixel-level reconstruction errors between each model output and the ground truth, then rank the pixel errors from small to large, divide them evenly into five intervals based on the ranking, and aggregate them.
In Figure~\ref{002_ob}(b), we discover that although Model(4) has three more convolution blocks than Model(1), it only achieves 17 and 10 error reductions in the 0$\sim$40 interval compared to Model(1) for the Middlebury and Vimeo90K datasets.
However, Model(4) achieves a decrease error of 346 and 251 in the 80$\sim$100 interval for these two datasets.
Inspired by this observation, we should allocate most computing resources to pixel locations in the 80$\sim$100 interval, since the VFI for these locations can be significantly improved with more computational resources.

Figure~\ref{002_ob}(d-e) illustrates the difference between Model(4) and ground truth, Model(1) and ground truth.
It can be observed that both Model(1) and Model(4) can predict well for the static or simple movement regions, indicating redundant computations in these regions and intrinsic spatial sparse for VFI.
However, both models have trouble predicting large or complex movement regions, which are crucial for visual pleasure in the VFI task.
From a Bayesian perspective \cite{NIPS2017_2650d608,NEURIPS2021_88a19961}, the targeted pixels in the reconstructed challenging region have large uncertainty (variance) as shown in the Figure~\ref{002_ob}(f).
Based on the aforementioned observation and discussion, we propose using uncertainty-generated mask label to guide our VFI framework allocating more computational resources to these challenging region.

\begin{figure*}[t]
\centering
\includegraphics[width=\linewidth]{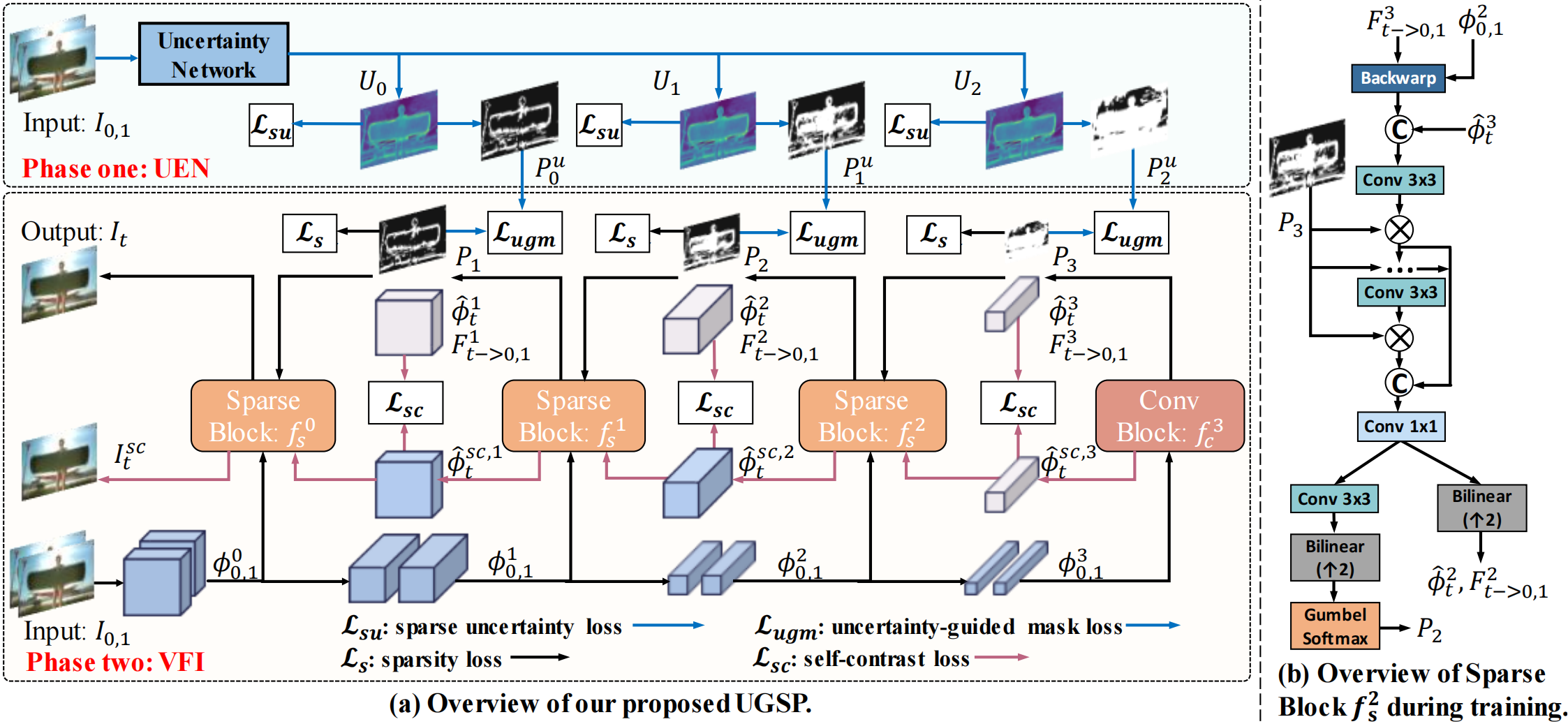}
\caption{Overview of UGSP. (a) In the first phase, we utilize $\mathcal{L}_{s u}$ to estimate uncertainty $U_k$ and generate pruning mask label $P_k$ in the levels 0,1, and 2. Then, in the second phase, we estimate pruning mask to skip the redundant computational under the $\mathcal{L}_{u g m}$ guide. In addition, we propose $\mathcal{L}_{s c}$ to increase the estimation ability of our UGSP by using the interpolated features generated by the auxiliary non-pruning branch. $\mathcal{L}_{s}$ controls the pruning degree of UGSP. (b) Detail of the Sparse Block $f_s^2$ during training.}
\label{003_method_model}
\end{figure*}
\subsection{Overview of the Proposed Framework}
\label{sec:overview}

As illustrated in Figure~\ref{003_method_model}, our UGSP consists of two training phases.
In the first phase, we train an uncertainty estimation network (UEN). As shown in the upper part of Figure~\ref{003_method_model}(a), UEN can predict the uncertainty (variance) field $(U_0, U_1, U_2)$ for the unknown intermediate frame.
The uncertainty map is then used to generate pruning masks $(P_0^u, P_1^u, P_2^u$) as a guide that supervises the second phase's spatial pruning mask $(P_1, P_2, P_3$) estimation.
In the second phase, spatial pruning masks at higher resolution scales are estimated from lower resolution scales, as shown in the bottom part of Figure~\ref{003_method_model}(a). Then, we use the pruning masks to skip redundant computations using sparse convolution in our VFI network.

In this section, we detail the structure of UEN and VFI networks.
The backbone network structure in UEN is identical to that of the VFI network, but UEN removes the branch for estimating the pruning mask and adds a branch for uncertainty estimation. We provide its structural details in the Appendix~\ref{appendix:ugsp}.
Our VFI network is designed in a coarse-to-fine manner similarly to most VFI models \cite{Kong_2022_CVPR,huang2022rife, Park_2021_ICCV,Lee_2020_CVPR}.

\begin{figure}[t]
\centering
\includegraphics[width=0.9\linewidth]{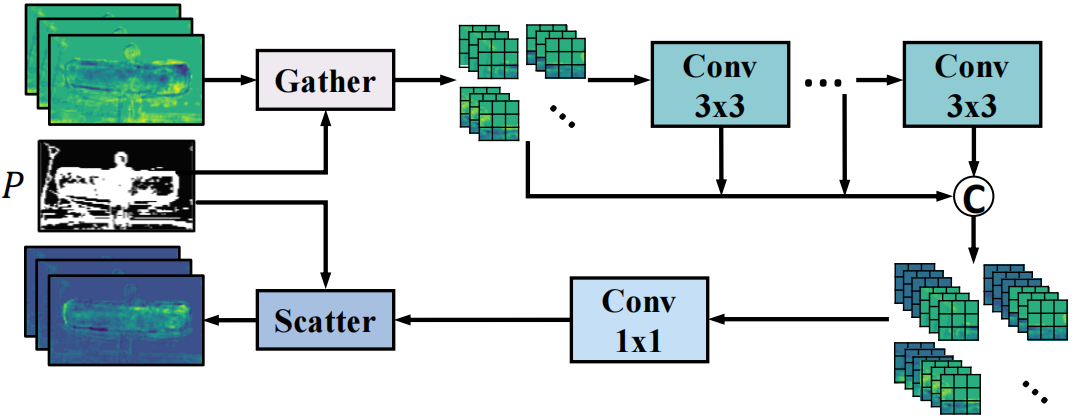}
\caption{ The process of the sparse convolution during inference.
}
\label{004_method_sparse}
\end{figure}

Specifically, as displayed in Figure~\ref{003_method_model}(a), it first downsamples two input frames $I_0$ and $I_1$ four times using a block of two 3×3 convolutions with strides 2 and 1 to obtain four level features $\phi_{0,1}^k, k \in\{0,1,2,3\}$.
Then, we predict the flow field $F_{t \rightarrow 0,1}^k$ and intermediate feature $\hat{\phi}_t^k$ for each scale level.
At the largest three scales, we use Sparse Block $(f_s^0, f_s^1, f_s^2)$ to skip redundant computations based on the spatial pruning mask $(P_1, P_2, P_3)$ which is estimated by Sparse Block $(f_s^1, f_s^2)$ and Conv Block $f_c^3$, respectively.
The details of Sparse Block $(f_s^0, f_s^1)$ and Conv Block $f_c^3$ is described in the Appendix~\ref{appendix:network}.
We provide the overview of the Sparse Block $f_s^2$ during training in Figure~\ref{003_method_model}(b).
We can see that we first concatenate the feature $\hat{\phi}_{t}^3$ from the Conv Block and the aligned feature produced by back warping using flow field $F_{t \rightarrow 0,1}^3$ and feature $\phi_{0,1}^2$.
Then we input it into five consecutive $3\times3$ convolutions and one $1\times1$ convolution to refine the feature. $P_3$ is used to skip the redundant computations in the $3\times3$ convolutions.
Finally, we generate flow field $F_{t \rightarrow 0,1}^2$, feature $\phi_{0,1}^2$ by using one bilinear up-sampling and generate pruning mask $P_2$ by using one $3\times3$ convolutions, one bilinear up-sampling and a Gumbel softmax.
$(\hat{\phi}_{t}^2, F_{t \rightarrow 0,1}^2, P_2)$ are used as input for next Sparse Block $f_s^1$.

\begin{figure*}[t]
\centering
\includegraphics[width=\linewidth]{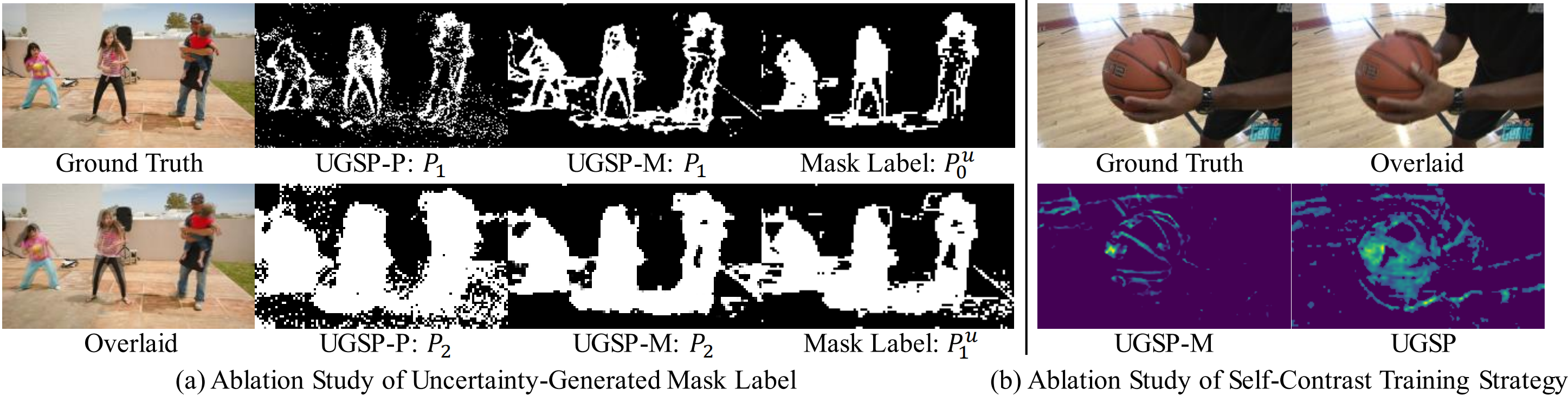}
\caption{ Ablation study. (a) Effectiveness of uncertainty-guided mask loss $\mathcal{L}_{ugm}$. $P_1$ and $P_2$ are the estimated pruning mask in VFI network during training, and $P_0^u$ and $P_1^u$ are uncertainty-generated mask label in UEN. (b) Effectiveness of self-contrast strategy. 
 UGSP-C use both the uncertainty-guide mask loss $\mathcal{L}_{u g m}$, and self-contrast loss $\mathcal{L}_{sc}$, while  UGSP-M only use uncertainty-guide mask loss $\mathcal{L}_{u g m}$.
}
\label{005_ablation_ugm_sc}
\end{figure*}

During training, as shown in Figure~\ref{003_method_model}(b), we achieve sparse convolution by using dotting product after each $3\times3$ convolutions.
The dotting product enables the backpropagation of gradients at all locations.
During inference, as shown in Figure~\ref{004_method_sparse}, we only allocate convolution operation on the important areas which we can index from the estimated pruning mask $P$. Specifically, we gather the select location according to the pruning mask, then we apply the $3\times3$ and $1\times1$ convolutions on them. After the $1\times1$ convolution, we scatter the feature back.
To control the pruning degree of UGSP, we utilize a sparse loss $\mathcal{L}_{s}$ on $P_k$:
\begin{equation}
 \begin{aligned}
\mathcal{L}_{s}= \left\| \frac{1}{\sum_{k=1}^{3} (H_k \times W_k)} (\sum_{k=1}^{3}\sum_{h=1}^{H_k}\sum_{w=1}^{W_k} P_{k,h,w}) - S_t  \right\|_{1},
\label{equ:s}
\end{aligned}
\end{equation}
where $H_k$ and $W_k$ is the height and width of the $P_k$, and $S_t$ is the target sparsity to control the FLOPs of our UGSP framework.

Similar to other VFI methods \cite{Niklaus_2018_CVPR,Niklaus_2020_CVPR,huang2022rife}, we use the image reconstruction loss to measure the difference between an interpolated frame $I_t$ and its corresponding ground truth $I_{gt}$:
\begin{equation}
 \begin{aligned}
\mathcal{L}_{rec}=  \left\| I_t - I_t^{gt}  \right\|_{1}.
\label{equ:u}
\end{aligned}
\end{equation}
Here, we use $\mathcal{L}_{1}$ between two laplacian pyramid representations of the output of the VFI network $I_t$ and the ground truth $I_t^{gt}$.

\subsection{Uncertainty-Guided Mask Prediction}
\label{sec:ugm}

In the previous study, large uncertainty can be used to identify semantically
and visually challenging pixels, such as object boundaries for semantic segmentation tasks \cite{NIPS2017_2650d608} or texture and edge pixels in super-resolution tasks \cite{NEURIPS2021_88a19961}.
In our VFI task, complex and large movement areas display large uncertainty, such as the ball movement areas in Figure~\ref{001_intro}(b).
Therefore, we propose estimating the pixel-by-pixel uncertainty field of the intermediate frame and using the uncertainty to guide the prediction of the pruning.
Following previous works \cite{NIPS2017_2650d608,NEURIPS2021_88a19961}, we utilize the sparse uncertainty loss $\mathcal{L}_{s u}$ to estimate the uncertainty field as follows:
\begin{equation}
 \begin{aligned}
\mathcal{L}_{s u}= \exp \left(-{U}\right)\left\|{I}_{t}-f\left({I}_{0},{I}_{1}\right)\right\|_{1}+2 {U},
\label{equ:su}
\end{aligned}
\end{equation}
where $f\left({I}_{0},{I}_{1}\right)$ and $U$ denote the learned mean and uncertainty (variance) of the intermediate frame, respectively.

As shown in Figure~\ref{003_method_model}, we first train a UEN to estimate the uncertainty field and then explicitly guide the spatial pruning mask estimation using the uncertainty in our VFI network.
Specifically, we use $\mathcal{L}_{s u}$ to estimate the variance $(U_0, U_1, U_2)$ for each scale, and the resolution of $U_k$ is the same as the input frame $I_{0,1}$. We provide the structural details of UEN in the Appendix~\ref{appendix:network}.
Then we generate the pruning mask label based on the uncertainty for 0,1 and 2 levels of the VFI network, which can be expressed as:
\begin{equation}
\begin{aligned}
T_k&=\operatorname{sort}\left(U_{k-1}\right)\left[\alpha_k\% \times H \times W \right], \\
P_{k, i, j}^u&= \begin{cases}1, & U_{k-1, i, j} > T_k \\
0, & \text { otherwise, }\end{cases}
\end{aligned}
\label{equ:label}
\end{equation}
where $k \in (1,2,3)$.  In Equation~(\ref{equ:label}), we first sort the value in $U_k$ from small to large and then assign threshold $T_k$ to the $\alpha_k\%$ smallest value of $U_{k-1}$. $H$ and $W$ are the height and width of $I_{0,1}$. Then we generate the pruning mask label $P_k^u$ with the ($i,j$) location set to 1 if $U_{k-1,i,j}$ is larger than the threshold $T_k$ and 0 otherwise. 
Value 0 in $P_k$ indicates the location where convolution will be skipped.
Finally, an uncertainty-guide mask loss $\mathcal{L}_{u g m}$ is imposed on the prediction of pruning mask $P_k$ in the VFI network to properly skip redundant computation as follow:
\begin{equation}
 \begin{aligned}
\mathcal{L}_{u g m}=  \left\| \left({P}_{k}^{u}\right)\downarrow_{2^{k+1}}-{P}_{k+1} \right\|_{1},
\label{equ:ugm}
\end{aligned}
\end{equation}
where ${P}_{k}^{u}$ and ${P}_{k+1} (k=0,1,2)$ stand for the guided pruning mask from the UEN and the estimated pruning mask from VFI network. $\downarrow_{2^{k+1}}$ denotes that we down-sample the ${P}_{k}^{u}$ by a factor of $2^{k+1}$ to keep the resolution consistently.


\begin{table*}[t]
    \centering
    \resizebox{\linewidth}{!}
{
    \begin{tabular}{lcccllcllcllc}
    \toprule
    \multirow{2}{*}{ Model} & \multirow{2}{*}{Pruning}  & \multirow{2}{*}{$\mathcal{L}_{ugm}$} & \multirow{2}{*}{$\mathcal{L}_{sc}$} & \multicolumn{3}{c}{Vimeo90K} & \multicolumn{3}{c}{UCF101} & \multicolumn{3}{c}{MiddleBury} \\
    \cmidrule(r){5-7} \cmidrule(r){8-10} \cmidrule(r){11-13} & & & & Time(s) & FLOPs(G) & PSNR & Time(s) & FLOPs(G) & PSNR & Time(s) & FLOPs(G) & PSNR \\
    \hline
    Baseline & - & - & - & 1.16 & 31.7 & 35.74 & 0.70 & 18.1 & 35.30 & 2.99 & 85.0 & 37.47 \\
    UGSP-P & \Checkmark & - & - & 0.44(-62\%) & 15.7(-50\%) & 35.45 & 0.26(-63\%) & 9.0(-50\%) & 35.29 & 1.12(-63\%) & 38.3(-55\%) & 36.48  \\
    UGSP-C & \Checkmark & - & \Checkmark & 0.47(-59\%) & 16.3(-49\%) & 35.61 & 0.25(-64\%) & 8.6(-52\%) & 35.28 & 1.10(-63\%) & 39.8(-53\%) & 36.90  \\
    UGSP-M & \Checkmark &  \Checkmark & - & 0.45(-61\%) & 15.5(-51\%) & \textbf{\textcolor{red}{35.63}} & 0.26(-63\%) & 8.7(-52\%) & 35.30 & 1.14(-62\%) & 40.7(-52\%) & 36.83 \\
    UGSP & \Checkmark &  \Checkmark & \Checkmark  & 0.45(-61\%) & 15.6(-51\%) & 35.62 & 0.26(-63\%) & 8.6(-52\%) & \textbf{\textcolor{red}{35.31}} & 1.10(-63\%) & 39.3(-54\%) & \textbf{\textcolor{red}{36.96}} \\
    \hline
    Baseline & - & - & - & 1.16 & 31.7 & 35.74 & 0.70 & 18.1 & 35.30 & 2.99 & 85.0 & 37.47 \\
    UGSP-large & \Checkmark & \Checkmark & \Checkmark & \textbf{\textcolor{red}{0.59(-49\%)}} & \textbf{\textcolor{red}{21.0(-34\%)}} & 35.72 & \textbf{\textcolor{red}{0.36(-49\%)}} & \textbf{\textcolor{red}{12.4(-31\%)}} & 35.31 & \textbf{\textcolor{red}{1.58(-47\%)}} & \textbf{\textcolor{red}{59.2(-30\%)}} & 37.46  \\
    \hline
    \end{tabular}
}
    \caption{ Quantitative comparison of the ablation study. We increase $S_t$ in sparse loss $L_s$ to train and achieve UGSP-large.  
    Time and FLOPs denote inference time and FLOPs percentage reduction compared to `Baseline'. While maintaining the performance, UGSP-large reduces 34\%/30\% FLOPs and 49\%/47\% Time on the Vimeo90K/MiddleBury datasets and UGSP reduces 52\% FLOPs and 63\% Time on the UCF101 dataset.
    }
    \label{tab:ablation-loss}
\end{table*}

\subsection{Self-Contrast Training Strategy}
\label{sec:sc}
To further enhance the estimation capacity of the model, we use the auxiliary non-pruning branch, which does not utilize pruning mask to skip computation and output the intermediate frame $I_t^{sc}$ and feature $\hat{\phi}_{t}^{sc,k}$. The auxiliary non-pruning branch shares the same weight with UGSP in each level. Then we propose a self-contrast loss $\mathcal{L}_{s c}$, which can be expressed as:
\begin{equation}
 \begin{aligned}
\mathcal{L}_{sc}= \mathcal{L}_{rec} \left(I_t^{sc}, I_t^{gt}  \right) + \sum_{i=1}^{3} \mathcal{L}_{cen} \left(\hat{\phi}_{t}^{sc,k}, \hat{\phi}_{t}^{k}  \right) ,
\label{equ:sc}
\end{aligned}
\end{equation}
where $I_t^{sc}$ and $\hat{\phi}_{t}^{sc,k}$ are the intermediate frame and features outputted from the auxiliary non-pruning branch. $\mathcal{L}_{cen}$ is the census loss \cite{Meister_Hur_Roth_2018}, which computes the soft Hamming distance for the intermediate features.
$\mathcal{L}_{s c}$ has two benefits for UGSP training. 
First, as UGSP skips the redundant computation areas by multiplying the features with pruning mask during training, the backward gradient propagation in the skipped areas is close to zero. To address this problem, the first part of $\mathcal{L}_{sc}$ applies $\mathcal{L}_{rec}$ to $I_t^{sc}$ and $I_t^{gt}$, producing a non-zero gradient in these regions. Second, analogous to the idea of knowledge distillation \cite{44873}, the intermediate feature $\hat{\phi}_{t}^{sc,k}$ generated from the model without pruning (teacher), can be regarded as a soft label that transfers knowledge to facilitate the performance of UGSP (student) by comparing the difference with $\hat{\phi}_{t}^{k}$.

\subsection{Overall Loss Function}
\label{sec:loss}
As described in Section~\ref{sec:ugm}, the loss function is sparse uncertainty loss $\mathcal{L}_{s u}$ (Equation~(\ref{equ:su})) for level 0, 1, and 2 in the first phase.
For the second phase, we summarize the loss described above as follows:
\begin{equation}
        \mathcal{L}_{{overall }}=\mathcal{L}_{{r e c }} +\lambda_{s}\mathcal{L}_{s} +\lambda_{u g m} \mathcal{L}_{u g m}+\lambda_{{s c}} \mathcal{L}_{{s c}}.
\label{equ:overall}
\end{equation}
$\lambda_{s}$, $\lambda_{u g m}$, and $\lambda_{{s c}}$ are the weight for $\mathcal{L}_{s}$, $\mathcal{L}_{u g m}$, and $\mathcal{L}_{{s c}}$.

\subsection{Relationships with Related Work}
\label{sec:rrw}
Many state-of-the-art efficient VFI methods, such as RIFE \cite{huang2022rife} and IFRnet \cite{Kong_2022_CVPR}, aim to design novel losses or apply the knowledge distillation strategy. By contrast, we propose another new strategy to achieve efficient VFI by designing a spatial pruning-friendly architecture and specific spatial pruning loss functions.
What is more, our architecture UGSP is compatible with previous methods.
Therefore, we also implement the flow distillation method and geometry consistency loss of IFRnet into our UGSP as UGSP-\textit{distill} and implement the priveleged distillation and refinement network of RIFE into UGSP as UGSP-\textit{refine}. We provide the implementation details in the Appendix~\ref{appendix:ugsp}.
The compared experiments in Section~\ref{sec:compare} shows that UGSP-\textit{distill} and UGSP-\textit{refine} can further improve the performance.

\section{Experiments}

This section begins by introducing benchmarks and evaluation metrics. Then, we analyze our UGSD framework to demonstrate the proposed design. Finally, we quantitatively and qualitatively compare UGSP with state-of-the-arts on the benchmark, and show the high compatibility of our framework. We introduce the implementation detail in the Appendix~\ref{appendix:implementation}.


\subsection{Benchmarks and Evaluation Metrics}
Our UGSP is trained on the Vimeo90K training dataset and evaluated on the Vimeo90K testing datasets, UCF 101, and Middlebury Other datasets.

\noindent
\textbf{Vimeo90k \cite{vimeo}:} A widely used dataset that has 51312 and 3782 triplets of size 256×448 for training and testing.

\noindent
\textbf{UCF101 \cite{ucf101}:} The DVF-selected \cite{Liu_2017_ICCV} test set having 379 triplets with a resolution of 256×256, is used to evaluate our algorithm.

\noindent
\textbf{Middlebury \cite{Baker2011}:} It is a widely dataset used for optical flow and VFI tasks. Other set is used for testing, and its image resolution is around 640×480.

We measure the peak signal-to-noise ratio (PSNR) for quantitative evaluation.
Each method traverses the three testing datasets on an Intel I5-10400K CPU to measure the CPU inference speed.
Additionally, we calculate floating point operations (FLOPs) to determine the computational complexity for each datasets.

\subsection{Method Analysis}
\label{sec:abaltion_study}

We conducted several experiments on the Vimeo90K, UCF101, and MiddleBury testing datasets to verify the effectiveness of the proposed approaches. 
Only $\mathcal{L}_{1}$ distance image reconstruction loss $\mathcal{L}_{r e c}$ (Equation~\ref{equ:u}) is used to train the `Baseline' network, and the result is shown in Table~\ref{tab:ablation-loss}. 
We adjust sparsity loss $\mathcal{L}_{s}$ to ensure that FLOPs is nearly equal across all variant models.

\noindent
\textbf{Ablation of Uncertainty-Generated Mask Label $(\mathcal{L}_{u g m})$.}
We first conduct experiments to demonstrate the effectiveness of the uncertainty-generated pruning mask.
We prune the baseline model using sparsity loss $\mathcal{L}_{s}$ but remove our designed loss ($\mathcal{L}_{ugm}$ and $\mathcal{L}_{sc}$), and name it UGSP-P.
Then we develop UGSP-M, which uses the uncertainty-generated mask label from UEN to guide the estimation of pruning mask in the VFI network through $\mathcal{L}_{ugm}$.
Table~\ref{tab:ablation-loss} shows that UGSP-M has superior PSNR than UGSP-P by 0.18dB and 0.35dB on the Vimeo90K and MiddleBury datasets, respectively.

\begin{figure}[t]
\centering
\includegraphics[width=0.9\linewidth]{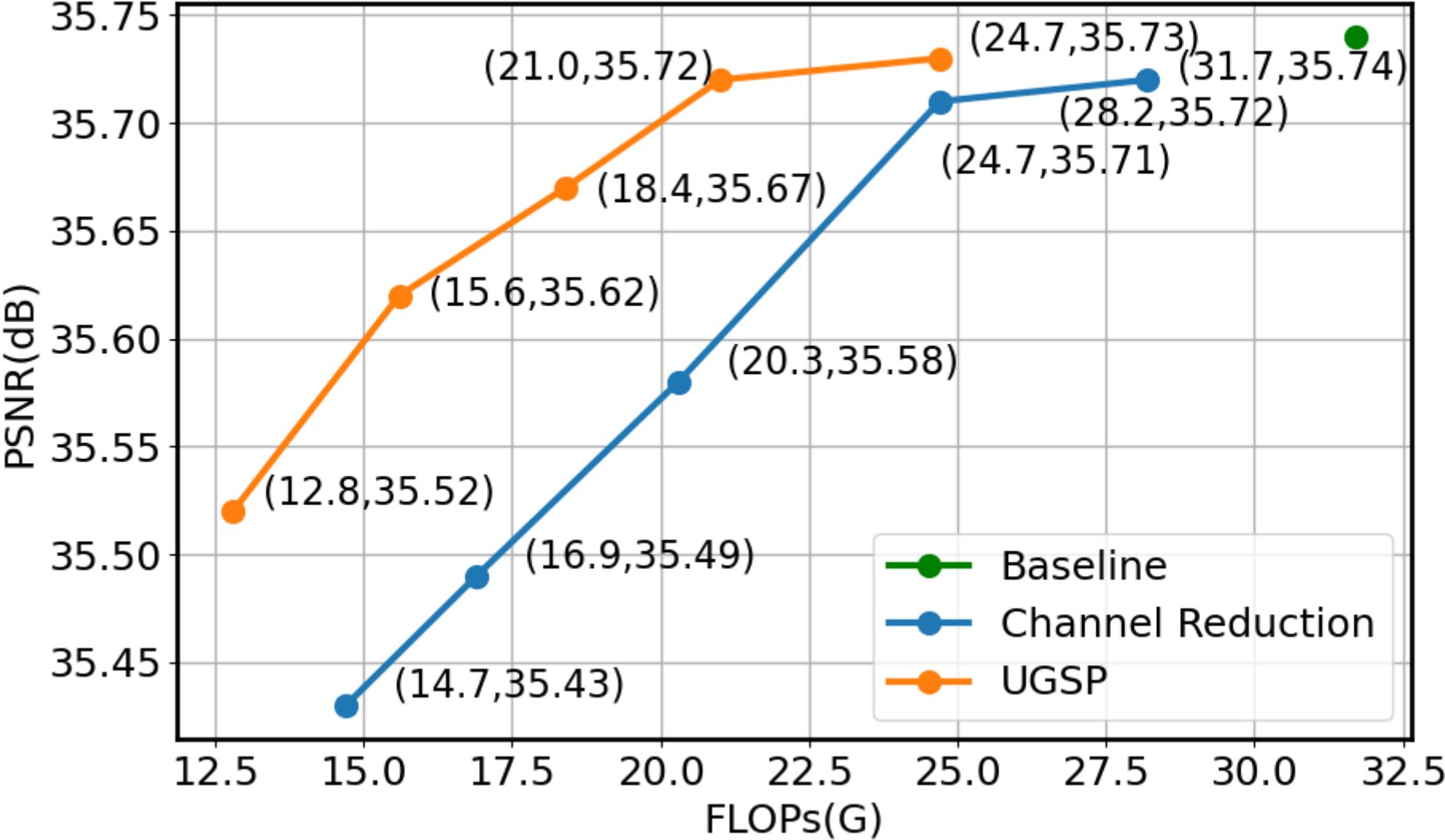}
\caption{ Analysis of the controllable computational complexity on the Vimeo90K dataset. 
}
\label{006_ablation_backbone}
\end{figure}

\begin{figure}[t]
\centering
\includegraphics[width=0.95\linewidth]{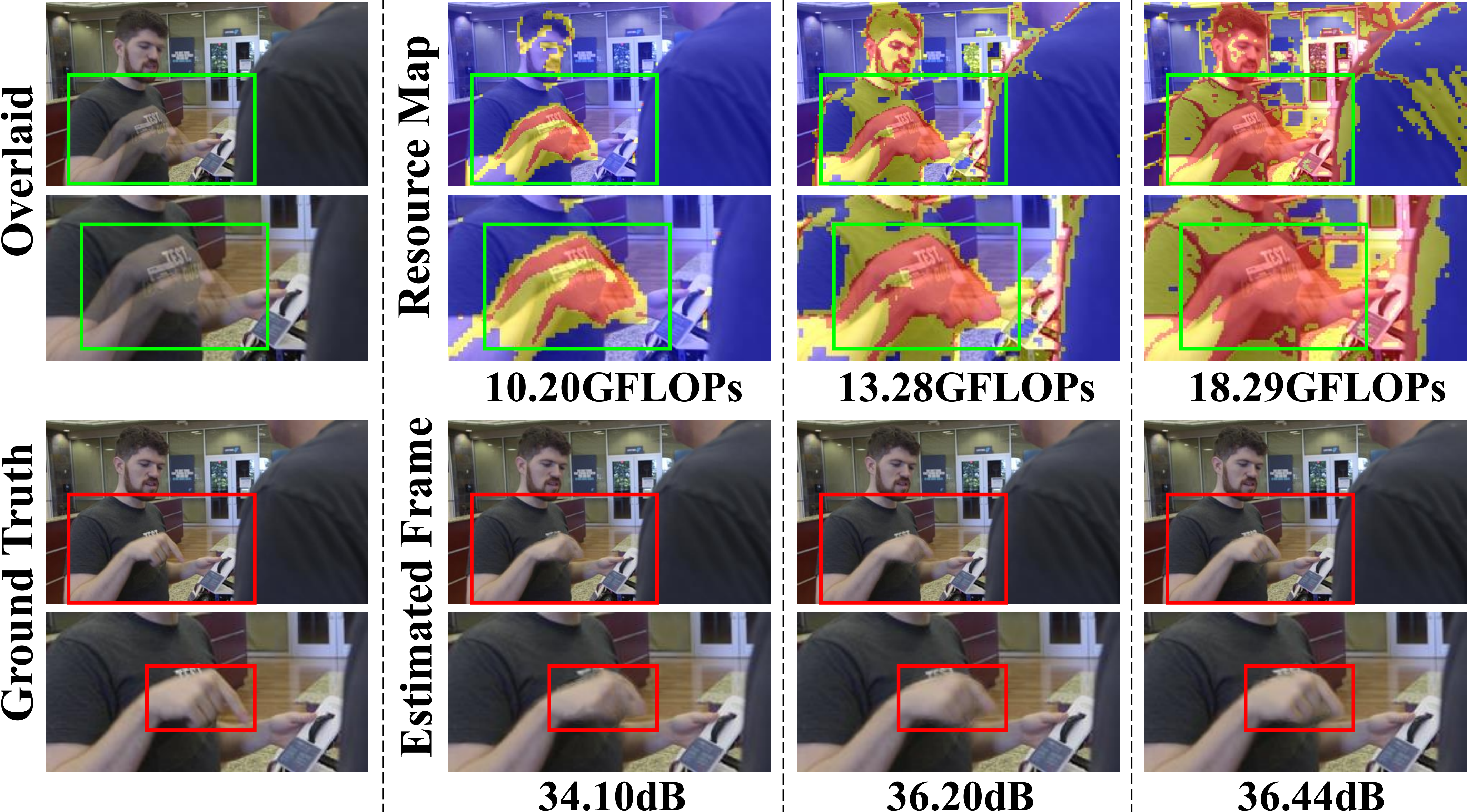}
\caption{ 
The effects of estimated pruning mask. In the first and second row, the red region denotes more computational resources allocated according to the pruing mask, while the blue area indicates less allocation.
}
\label{006_ablation_rebuttal}
\end{figure}

Additionally, we conducted a qualitative ablation study, as shown in Figure~\ref{005_ablation_ugm_sc}(a). Figure~\ref{005_ablation_ugm_sc}(a) presents the estimated pruning masks ($P_1, P_2$) and uncertainty-generated mask labels ($P_0^u, P_1^u$).
We observe that the white areas requiring convolution are more concentrated in UGSP-M and `Mask Label' than UGSP-P. This is advantageous to large and complex motion estimation because more contextual information of motion is included. In addition, the scatter and small white regions in UGSP-P increase the number of unnecessary calculations since the receptive field is too limited to capture the motion.
In short, using the uncertainty-generated mask label, the performance of our model can significantly enhance.

\begin{table*}[tb]
    \centering
    \resizebox{.8\linewidth}{!}
{
    \begin{tabular}{lccccccccc}
    \toprule
   \multirow{2}{*}{ Model} &  \multicolumn{3}{c}{Vimeo90K} & \multicolumn{3}{c}{UCF101} &  \multicolumn{3}{c}{Middlebury} \\
     \cmidrule(r){2-4} \cmidrule(r){5-7} \cmidrule(r){8-10} & Time(s) & FLOPs(G) & PSNR & Time(s) & FLOPs(G) & PSNR & Time(s) &  FLOPs(G) & PSNR \\
    \hline
    SepConv \cite{Niklaus_2017_ICCV} & - & 44.00 & 33.79 & - & 25.14 & 34.78 & - & 117.87 & 35.85 \\
    CAIN \cite{Choi_Kim_Han_Xu_Lee_2020} & 2.80 & 171.98 & 34.65 & 1.38 & 85.99 & 34.98 & 6.72 & 429.95 & 35.08 \\
    AdaCof \cite{Lee_2020_CVPR} & - & 43.73 & 34.38 & - & 24.99 & 35.20 & - & 117.13 & 35.74  \\
    EDSC \cite{9501506} & - & 31.55 & 34.84 & - & 18.03 & 35.13 & - & 84.52 & 36.80  \\
    BMBC \cite{10.1007/978-3-030-58568-6_7} & 20.51 & 305.33 & 35.01 & 12.35 & 174.47 & 35.15 & 47.88 & 817.84 & 36.79 \\
    CDFI \cite{Ding_2021_CVPR} & - & 100.26 & 35.17 & - & 57.29 & 35.21 & - & 268.56 & 37.14\\
    IFRnet \cite{Kong_2022_CVPR} & 0.62 & \textcolor{blue}{\underline{15.19}} & 35.59 & 0.38 & 8.68 & 35.28 & 1.56 & 40.68 &  37.21  \\
    RIFE \cite{huang2022rife} & 0.66 & 20.45 & \textcolor{blue}{\underline{35.62}} & 0.39 & 11.68 & 35.28  & 1.70 & 54.77 & \textcolor{blue}{\underline{37.29}}  \\
    \hline
    UGSP & \textcolor{blue}{\underline{0.46}} & 15.60 & \textcolor{blue}{\underline{35.62}} & \textcolor{blue}{\underline{0.26}} & \textcolor{blue}{\underline{8.54}} & \textcolor{blue}{\underline{35.31}} & \textcolor{blue}{\underline{1.10}} & \textbf{\textcolor{red}{39.29}}  & 36.96 \\
    UGSP-\textit{distill} & \textbf{\textcolor{red}{0.43}} & \textbf{\textcolor{red}{14.72}} & \textbf{\textcolor{red}{35.65}} & \textbf{\textcolor{red}{0.23}} & \textbf{\textcolor{red}{8.00}} & 35.27 & \textbf{\textcolor{red}{1.04}} & \textcolor{blue}{\underline{40.48}} & 36.98  \\
    UGSP-\textit{refine} & 0.59 & 19.73 & \textcolor{blue}{\underline{35.62}} & 0.35 & 11.25 & \textbf{\textcolor{red}{35.33}} & 1.52 & 53.12 & \textbf{\textcolor{red}{37.31}}  \\



    
    \hline
    \end{tabular}
}
    \caption{ 
 Quantitative comparison.
 For each item, best results are \textbf{\textcolor{red}{bold}}, and the second best is \textcolor{blue}{\underline{underlined}}. The symbol '-' indicates that the method is not testable on the CPU due to the absence of its implementation in the CPU-oriented version.
}
    \label{tab:compare}
\end{table*}

\noindent
\textbf{Ablation of $\mathcal{L}_{sc}$.}
As shown in Table~\ref{tab:ablation-loss}, using our $\mathcal{L}_{sc}$, our UGSP-C performs better than UGSP-P, and our UGSP can increase PSNR by 0.13dB compared to UGSP-M on the MiddleBury dataset. UGSP achieves a reduction of 52\% in FLOPs and 63\% in inference time while maintaining performance on the UCF101 dataset. Additionally, UGSP-large reduces FLOPs by 34\%/30\% and inference time by 49\%/47\%, while keeping  a comparable PSNR performance to the 'Baseline' model (35.72 vs. 35.74 for Vimeo90K, and 37.46 vs. 37.47 for MiddleBury).
Our proposed self-contrast training strategy ($\mathcal{L}_{sc}$) guides UGSP training using the output of non-pruning branch.
As shown in Figure~\ref{005_ablation_ugm_sc}(b), the intermediate feature of UGSP reveals high activation in the motion areas, which is vital to a pleasing visual experience.

    

\noindent
\textbf{Controllable Computational Complexity.}
We can easily control the FLOPs for UGSP by setting different target sparsity hyperparameter $S_t$ in the $\mathcal{L}_{s}$ (Equation~\ref{equ:s}) to train the model.
As presented in Figure~\ref{006_ablation_backbone}, we change $S_t$ to achieve UGSP with different FLOPs and compare this to the `Channel Reduction'. `Channel Reduction' indicates that we scale feature channels in `Baseline' to reduce FLOPs.
We find that UGSP has superior PSNR performance than `Channel Reduction', since UGSP allocates more computation resources to challenging areas, which can significantly improve results as discussed in Section~\ref{sec:ob}.

 We provide a qualitative understanding of the effects of estimated pruing mask in Figure~\ref{006_ablation_rebuttal}. It shows that when the computational complexity around the hands (red regions) increases, it predicts hand motion more accurately. In addition, it can be observed that the allocation of computational resources is concentrated in the regions associated with motion, which is crucial to obtain pleasing visual results.



\begin{figure*}[t]
\centering
\includegraphics[width=0.95\linewidth]{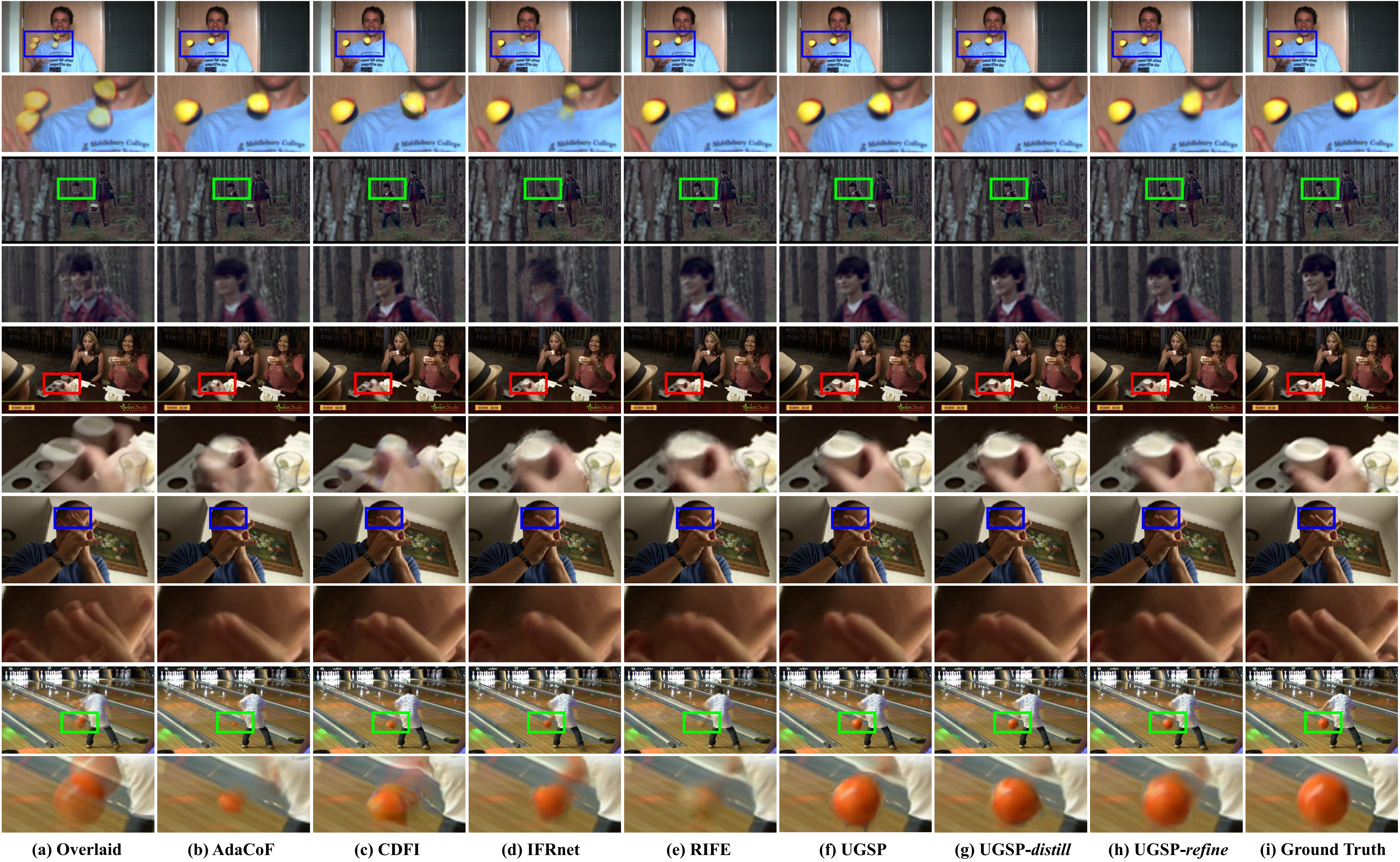}
\caption{ Qualitative comparisons. The first example comes from the Middlebury, and other examples are from the Vimeo90K.  
}
\label{006_compare}
\end{figure*}

\subsection{Comparison with the State-of-the-Arts}
\label{sec:compare}
Our UGSP, UGSP-\textit{distill}, and UGSP-\textit{refine} are compared to 8 state-of-the-art methods, including SepConv \cite{Niklaus_2017_ICCV}, CAIN \cite{Choi_Kim_Han_Xu_Lee_2020}, AdaCof \cite{Lee_2020_CVPR}, EDSC \cite{9501506}, BMBC \cite{10.1007/978-3-030-58568-6_7}, CDFI \cite{Ding_2021_CVPR}, RIFE \cite{huang2022rife}, and IFRnet \cite{Kong_2022_CVPR}.

\noindent
\textbf{Compatibility of UGSP.}
Other VFI methods can be applied to UGSP as we discussed in Section~\ref{sec:rrw}.
We implement the flow distillation and geometry consistency loss of IFRnet \cite{Kong_2022_CVPR} into UGSP as UGSP-\textit{distill}, and we add refinement network and priveleged distillation strategy of RIFE \cite{huang2022rife} into UGSP as UGSP-\textit{refine}. As shown in Table~\ref{tab:compare}, UGSP-\textit{distill} performs superior to UGSP and IFRnet by 0.03dB and 0.06dB PSNR but achieves 0.88G and 0.41G lower FLOPs on the Vimeo90K dataset.
Moreover, UGSP-\textit{distill} outperforms IFRnet in significantly reducing inference time.
The UGSP-\textit{refine} achieves the best PSNR on the UCF101 and MiddleBury compared to UGSP and RIFE with lower inference time.




\noindent
\textbf{Quantitative Results.}
Table~\ref{tab:compare} shows that our UGSP, UGSP-\textit{refine}, and UGSP-\textit{distill} outperform the state-of-the-art methods on most datasets.
For example, our UGSP-\textit{distill} achieves the best 35.65dB PSNR values for the Vimeo90K dataset with lower FLOPs.
In addition, despite other methods, such as RIFE, CDFI, and BMBC, having larger FLOPs than UGSP-\textit{refine}, our UGSP-\textit{refine} performs favorably against them on all benchmarks.
Furthermore, without combing our UGSP with RIFE or IFRnet, UGSP has already achieved the highest performance on Vimeo90K and UCF101 datasets with minimal FLOPs.

\noindent
\textbf{Qualitative Results.}
The visual results shown in Figure~\ref{006_compare} reveal that our framework can interpolate large and complex regions.
For example, our UGSP produces a sharp yellow ball in the first examples in Figure~\ref{006_compare}, and UGSP-\textit{distill} maintains the structure of the ball in the last example.
In short, as our approaches guide the computation reducing only in the simple and static regions, we can maintain the high-quality in challenging movement regions, which is important for the pleasing visual experience.

\section{Conclusion}
In this paper, we propose the UGSP framework for video frame interpolation that aims to skip redundant computation using uncertainty as a guide.
We observe that low uncertainty regions appear in the easy regions, such as small-scale motion, which can be skipped to increase inference speed.
Therefore, a UEN in our UGSP framework first estimates uncertainty. Then VFI network uses uncertainty to guide the pruning mask estimation through proposed uncertainty-guided mask loss, which increases a significant PSNR value compared to when uncertainty is not used.
In addition, by including the proposed self-contrast training strategy, our UGSP can effectively reduce computational cost while maintaining comparable performance to baseline.
UGSP achieves 34\%/52\%/30\% FLOPs and 49\%/49\%/47\% inference time compared to baseline, but maintains the PSNR performance on Vimeo90K/UCF101/MiddleBury datasets.
What is more, other efficient VFI network can be integrated into our UGSP. 
Extensive experiments demonstrate that our network also achieve state-of-the-art performance with large FLOPs reduction and a speedup on CPU devices.

\bibliographystyle{ACM-Reference-Format}
\balance
\bibliography{samples/egbib}


\begin{thebibliography}{54}


\ifx \showCODEN    \undefined \def \showCODEN     #1{\unskip}     \fi
\ifx \showDOI      \undefined \def \showDOI       #1{#1}\fi
\ifx \showISBNx    \undefined \def \showISBNx     #1{\unskip}     \fi
\ifx \showISBNxiii \undefined \def \showISBNxiii  #1{\unskip}     \fi
\ifx \showISSN     \undefined \def \showISSN      #1{\unskip}     \fi
\ifx \showLCCN     \undefined \def \showLCCN      #1{\unskip}     \fi
\ifx \shownote     \undefined \def \shownote      #1{#1}          \fi
\ifx \showarticletitle \undefined \def \showarticletitle #1{#1}   \fi
\ifx \showURL      \undefined \def \showURL       {\relax}        \fi
\providecommand\bibfield[2]{#2}
\providecommand\bibinfo[2]{#2}
\providecommand\natexlab[1]{#1}
\providecommand\showeprint[2][]{arXiv:#2}

\bibitem[Badrinarayanan et~al\mbox{.}(2017)]%
        {7803544}
\bibfield{author}{\bibinfo{person}{Vijay Badrinarayanan}, \bibinfo{person}{Alex
  Kendall}, {and} \bibinfo{person}{Roberto Cipolla}.}
  \bibinfo{year}{2017}\natexlab{}.
\newblock \showarticletitle{SegNet: A Deep Convolutional Encoder-Decoder
  Architecture for Image Segmentation}.
\newblock \bibinfo{journal}{\emph{IEEE Transactions on Pattern Analysis and
  Machine Intelligence}} \bibinfo{volume}{39}, \bibinfo{number}{12}
  (\bibinfo{year}{2017}), \bibinfo{pages}{2481--2495}.
\newblock


\bibitem[Baker et~al\mbox{.}(2011)]%
        {Baker2011}
\bibfield{author}{\bibinfo{person}{Simon Baker}, \bibinfo{person}{Daniel
  Scharstein}, \bibinfo{person}{J.~P. Lewis}, \bibinfo{person}{Stefan Roth},
  \bibinfo{person}{Michael~J. Black}, {and} \bibinfo{person}{Richard
  Szeliski}.} \bibinfo{year}{2011}\natexlab{}.
\newblock \showarticletitle{A Database and Evaluation Methodology for Optical
  Flow}.
\newblock \bibinfo{journal}{\emph{International Journal of Computer Vision}}
  \bibinfo{volume}{92}, \bibinfo{number}{1} (\bibinfo{year}{2011}),
  \bibinfo{pages}{1--31}.
\newblock


\bibitem[Baker et~al\mbox{.}(2007)]%
        {vimeo}
\bibfield{author}{\bibinfo{person}{S. Baker}, \bibinfo{person}{R. Szeliski},
  \bibinfo{person}{M.~J. Black}, \bibinfo{person}{D. Scharstein},
  \bibinfo{person}{J. Lewis}, {and} \bibinfo{person}{S. Roth}.}
  \bibinfo{year}{2007}\natexlab{}.
\newblock \showarticletitle{A Database and Evaluation Methodology for Optical
  Flow}. In \bibinfo{booktitle}{\emph{2007 11th IEEE International Conference
  on Computer Vision}}. \bibinfo{publisher}{IEEE Computer Society},
  \bibinfo{address}{Los Alamitos, CA, USA}, \bibinfo{pages}{1--8}.
\newblock


\bibitem[Bao et~al\mbox{.}(2019)]%
        {Bao_2019_CVPR}
\bibfield{author}{\bibinfo{person}{Wenbo Bao}, \bibinfo{person}{Wei-Sheng Lai},
  \bibinfo{person}{Chao Ma}, \bibinfo{person}{Xiaoyun Zhang},
  \bibinfo{person}{Zhiyong Gao}, {and} \bibinfo{person}{Ming-Hsuan Yang}.}
  \bibinfo{year}{2019}\natexlab{}.
\newblock \showarticletitle{Depth-Aware Video Frame Interpolation}. In
  \bibinfo{booktitle}{\emph{Proceedings of the IEEE/CVF Conference on Computer
  Vision and Pattern Recognition (CVPR)}}.
\newblock


\bibitem[Bao et~al\mbox{.}(2021)]%
        {8840983}
\bibfield{author}{\bibinfo{person}{Wenbo Bao}, \bibinfo{person}{Wei-Sheng Lai},
  \bibinfo{person}{Xiaoyun Zhang}, \bibinfo{person}{Zhiyong Gao}, {and}
  \bibinfo{person}{Ming-Hsuan Yang}.} \bibinfo{year}{2021}\natexlab{}.
\newblock \showarticletitle{MEMC-Net: Motion Estimation and Motion Compensation
  Driven Neural Network for Video Interpolation and Enhancement}.
\newblock \bibinfo{journal}{\emph{IEEE Transactions on Pattern Analysis and
  Machine Intelligence}} \bibinfo{volume}{43}, \bibinfo{number}{3}
  (\bibinfo{year}{2021}), \bibinfo{pages}{933--948}.
\newblock


\bibitem[Bolukbasi et~al\mbox{.}(2017)]%
        {pmlr-v70-bolukbasi17a}
\bibfield{author}{\bibinfo{person}{Tolga Bolukbasi}, \bibinfo{person}{Joseph
  Wang}, \bibinfo{person}{Ofer Dekel}, {and} \bibinfo{person}{Venkatesh
  Saligrama}.} \bibinfo{year}{2017}\natexlab{}.
\newblock \showarticletitle{Adaptive Neural Networks for Efficient Inference}.
  In \bibinfo{booktitle}{\emph{Proceedings of the 34th International Conference
  on Machine Learning}} \emph{(\bibinfo{series}{Proceedings of Machine Learning
  Research}, Vol.~\bibinfo{volume}{70})}. \bibinfo{pages}{527--536}.
\newblock


\bibitem[{Cai} et~al\mbox{.}(2022)]%
        {2022arXiv220304845C}
\bibfield{author}{\bibinfo{person}{Yuanhao {Cai}}, \bibinfo{person}{Jing
  {Lin}}, \bibinfo{person}{Xiaowan {Hu}}, \bibinfo{person}{Haoqian {Wang}},
  \bibinfo{person}{Xin {Yuan}}, \bibinfo{person}{Yulun {Zhang}},
  \bibinfo{person}{Radu {Timofte}}, {and} \bibinfo{person}{Luc {Van Gool}}.}
  \bibinfo{year}{2022}\natexlab{}.
\newblock \showarticletitle{{Coarse-to-Fine Sparse Transformer for
  Hyperspectral Image Reconstruction}}.
\newblock \bibinfo{journal}{\emph{arXiv e-prints}} (\bibinfo{date}{March}
  \bibinfo{year}{2022}).
\newblock


\bibitem[Chang et~al\mbox{.}(2020)]%
        {Chang_2020_CVPR}
\bibfield{author}{\bibinfo{person}{Jie Chang}, \bibinfo{person}{Zhonghao Lan},
  \bibinfo{person}{Changmao Cheng}, {and} \bibinfo{person}{Yichen Wei}.}
  \bibinfo{year}{2020}\natexlab{}.
\newblock \showarticletitle{Data Uncertainty Learning in Face Recognition}. In
  \bibinfo{booktitle}{\emph{Proceedings of the IEEE/CVF Conference on Computer
  Vision and Pattern Recognition (CVPR)}}.
\newblock


\bibitem[Cheng and Chen(2020)]%
        {Cheng_Chen_2020}
\bibfield{author}{\bibinfo{person}{Xianhang Cheng} {and}
  \bibinfo{person}{Zhenzhong Chen}.} \bibinfo{year}{2020}\natexlab{}.
\newblock \showarticletitle{Video Frame Interpolation via Deformable Separable
  Convolution}.
\newblock \bibinfo{journal}{\emph{Proceedings of the AAAI Conference on
  Artificial Intelligence}}  \bibinfo{volume}{34} (\bibinfo{date}{Apr.}
  \bibinfo{year}{2020}), \bibinfo{pages}{10607--10614}.
\newblock


\bibitem[Cheng and Chen(2022)]%
        {9501506}
\bibfield{author}{\bibinfo{person}{Xianhang Cheng} {and}
  \bibinfo{person}{Zhenzhong Chen}.} \bibinfo{year}{2022}\natexlab{}.
\newblock \showarticletitle{Multiple Video Frame Interpolation via Enhanced
  Deformable Separable Convolution}.
\newblock \bibinfo{journal}{\emph{IEEE Transactions on Pattern Analysis and
  Machine Intelligence}} \bibinfo{volume}{44}, \bibinfo{number}{10}
  (\bibinfo{year}{2022}), \bibinfo{pages}{7029--7045}.
\newblock


\bibitem[Chi et~al\mbox{.}(2020)]%
        {10.1007/978-3-030-58583-9_7}
\bibfield{author}{\bibinfo{person}{Zhixiang Chi}, \bibinfo{person}{Rasoul
  Mohammadi~Nasiri}, \bibinfo{person}{Zheng Liu}, \bibinfo{person}{Juwei Lu},
  \bibinfo{person}{Jin Tang}, {and} \bibinfo{person}{Konstantinos~N.
  Plataniotis}.} \bibinfo{year}{2020}\natexlab{}.
\newblock \showarticletitle{All at Once: Temporally Adaptive Multi-frame
  Interpolation with Advanced Motion Modeling}. In
  \bibinfo{booktitle}{\emph{Computer Vision -- ECCV 2020}}.
  \bibinfo{pages}{107--123}.
\newblock


\bibitem[Choi et~al\mbox{.}(2020)]%
        {Choi_Kim_Han_Xu_Lee_2020}
\bibfield{author}{\bibinfo{person}{Myungsub Choi}, \bibinfo{person}{Heewon
  Kim}, \bibinfo{person}{Bohyung Han}, \bibinfo{person}{Ning Xu}, {and}
  \bibinfo{person}{Kyoung~Mu Lee}.} \bibinfo{year}{2020}\natexlab{}.
\newblock \showarticletitle{Channel Attention Is All You Need for Video Frame
  Interpolation}.
\newblock \bibinfo{journal}{\emph{Proceedings of the AAAI Conference on
  Artificial Intelligence}} \bibinfo{volume}{34}, \bibinfo{number}{07}
  (\bibinfo{date}{Apr.} \bibinfo{year}{2020}), \bibinfo{pages}{10663--10671}.
\newblock


\bibitem[Choi et~al\mbox{.}(2021)]%
        {Choi_2021_ICCV}
\bibfield{author}{\bibinfo{person}{Myungsub Choi}, \bibinfo{person}{Suyoung
  Lee}, \bibinfo{person}{Heewon Kim}, {and} \bibinfo{person}{Kyoung~Mu Lee}.}
  \bibinfo{year}{2021}\natexlab{}.
\newblock \showarticletitle{Motion-Aware Dynamic Architecture for Efficient
  Frame Interpolation}. In \bibinfo{booktitle}{\emph{Proceedings of the
  IEEE/CVF International Conference on Computer Vision (ICCV)}}.
  \bibinfo{pages}{13839--13848}.
\newblock


\bibitem[Ding et~al\mbox{.}(2021)]%
        {Ding_2021_CVPR}
\bibfield{author}{\bibinfo{person}{Tianyu Ding}, \bibinfo{person}{Luming
  Liang}, \bibinfo{person}{Zhihui Zhu}, {and} \bibinfo{person}{Ilya Zharkov}.}
  \bibinfo{year}{2021}\natexlab{}.
\newblock \showarticletitle{CDFI: Compression-Driven Network Design for Frame
  Interpolation}. In \bibinfo{booktitle}{\emph{Proceedings of the IEEE/CVF
  Conference on Computer Vision and Pattern Recognition (CVPR)}}.
  \bibinfo{pages}{8001--8011}.
\newblock


\bibitem[Fang et~al\mbox{.}(2022)]%
        {tmp}
\bibfield{author}{\bibinfo{person}{Zhenxuan Fang}, \bibinfo{person}{Weisheng
  Dong}, \bibinfo{person}{Xin Li}, \bibinfo{person}{Jinjian Wu},
  \bibinfo{person}{Leida Li}, {and} \bibinfo{person}{Shi Guangming}.}
  \bibinfo{year}{2022}\natexlab{}.
\newblock \showarticletitle{Uncertainty Learning in Kernel Estimation for
  Multi-Stage Blind Image Super-Resolution}. In
  \bibinfo{booktitle}{\emph{Proceedings of the European Conference on Computer
  Vision (ECCV)}}.
\newblock


\bibitem[Habibian et~al\mbox{.}(2021)]%
        {Habibian_2021_CVPR}
\bibfield{author}{\bibinfo{person}{Amirhossein Habibian},
  \bibinfo{person}{Davide Abati}, \bibinfo{person}{Taco~S. Cohen}, {and}
  \bibinfo{person}{Babak~Ehteshami Bejnordi}.} \bibinfo{year}{2021}\natexlab{}.
\newblock \showarticletitle{Skip-Convolutions for Efficient Video Processing}.
  In \bibinfo{booktitle}{\emph{Proceedings of the IEEE/CVF Conference on
  Computer Vision and Pattern Recognition (CVPR)}}.
  \bibinfo{pages}{2695--2704}.
\newblock


\bibitem[Han et~al\mbox{.}(2022)]%
        {9560049}
\bibfield{author}{\bibinfo{person}{Yizeng Han}, \bibinfo{person}{Gao Huang},
  \bibinfo{person}{Shiji Song}, \bibinfo{person}{Le Yang},
  \bibinfo{person}{Honghui Wang}, {and} \bibinfo{person}{Yulin Wang}.}
  \bibinfo{year}{2022}\natexlab{}.
\newblock \showarticletitle{Dynamic Neural Networks: A Survey}.
\newblock \bibinfo{journal}{\emph{IEEE Transactions on Pattern Analysis and
  Machine Intelligence}} \bibinfo{volume}{44}, \bibinfo{number}{11}
  (\bibinfo{year}{2022}), \bibinfo{pages}{7436--7456}.
\newblock


\bibitem[Hinton et~al\mbox{.}(2015)]%
        {44873}
\bibfield{author}{\bibinfo{person}{Geoffrey Hinton}, \bibinfo{person}{Oriol
  Vinyals}, {and} \bibinfo{person}{Jeffrey Dean}.}
  \bibinfo{year}{2015}\natexlab{}.
\newblock \showarticletitle{Distilling the Knowledge in a Neural Network}. In
  \bibinfo{booktitle}{\emph{NIPS Deep Learning and Representation Learning
  Workshop}}.
\newblock
\urldef\tempurl%
\url{http://arxiv.org/abs/1503.02531}
\showURL{%
\tempurl}


\bibitem[Huang et~al\mbox{.}(2023)]%
        {10096540}
\bibfield{author}{\bibinfo{person}{Chenyu Huang}, \bibinfo{person}{Weimin Tan},
  \bibinfo{person}{Jiaxing Shi}, \bibinfo{person}{Zhen Xing}, {and}
  \bibinfo{person}{Bo Yan}.} \bibinfo{year}{2023}\natexlab{}.
\newblock \showarticletitle{Uncer2Natural: Uncertainty-Aware Unsupervised Image
  Denoising}. In \bibinfo{booktitle}{\emph{ICASSP 2023 - 2023 IEEE
  International Conference on Acoustics, Speech and Signal Processing
  (ICASSP)}}. \bibinfo{pages}{1--5}.
\newblock
\urldef\tempurl%
\url{https://doi.org/10.1109/ICASSP49357.2023.10096540}
\showDOI{\tempurl}


\bibitem[Huang et~al\mbox{.}(2018)]%
        {huang2018multi}
\bibfield{author}{\bibinfo{person}{Gao Huang}, \bibinfo{person}{Danlu Chen},
  \bibinfo{person}{Tianhong Li}, \bibinfo{person}{Felix Wu},
  \bibinfo{person}{Laurens van~der Maaten}, {and} \bibinfo{person}{Kilian~Q
  Weinberger}.} \bibinfo{year}{2018}\natexlab{}.
\newblock \showarticletitle{Multi-scale dense networks for resource efficient
  image classification}.
\newblock \bibinfo{journal}{\emph{ICLR}}.
\newblock


\bibitem[Huang et~al\mbox{.}(2022)]%
        {huang2022rife}
\bibfield{author}{\bibinfo{person}{Zhewei Huang}, \bibinfo{person}{Tianyuan
  Zhang}, \bibinfo{person}{Wen Heng}, \bibinfo{person}{Boxin Shi}, {and}
  \bibinfo{person}{Shuchang Zhou}.} \bibinfo{year}{2022}\natexlab{}.
\newblock \showarticletitle{Real-Time Intermediate Flow Estimation for Video
  Frame Interpolation}. In \bibinfo{booktitle}{\emph{Proceedings of the
  European Conference on Computer Vision (ECCV)}}.
\newblock


\bibitem[Jiang et~al\mbox{.}(2018)]%
        {Jiang_2018_CVPR}
\bibfield{author}{\bibinfo{person}{Huaizu Jiang}, \bibinfo{person}{Deqing Sun},
  \bibinfo{person}{Varun Jampani}, \bibinfo{person}{Ming-Hsuan Yang},
  \bibinfo{person}{Erik Learned-Miller}, {and} \bibinfo{person}{Jan Kautz}.}
  \bibinfo{year}{2018}\natexlab{}.
\newblock \showarticletitle{Super SloMo: High Quality Estimation of Multiple
  Intermediate Frames for Video Interpolation}. In
  \bibinfo{booktitle}{\emph{Proceedings of the IEEE Conference on Computer
  Vision and Pattern Recognition (CVPR)}}.
\newblock


\bibitem[Kendall and Gal(2017)]%
        {NIPS2017_2650d608}
\bibfield{author}{\bibinfo{person}{Alex Kendall} {and} \bibinfo{person}{Yarin
  Gal}.} \bibinfo{year}{2017}\natexlab{}.
\newblock \showarticletitle{What Uncertainties Do We Need in Bayesian Deep
  Learning for Computer Vision?}. In \bibinfo{booktitle}{\emph{Advances in
  Neural Information Processing Systems}}, Vol.~\bibinfo{volume}{30}.
  \bibinfo{publisher}{Curran Associates, Inc.}
\newblock


\bibitem[Kim et~al\mbox{.}(2020)]%
        {Kim_Oh_Kim_2020}
\bibfield{author}{\bibinfo{person}{Soo~Ye Kim}, \bibinfo{person}{Jihyong Oh},
  {and} \bibinfo{person}{Munchurl Kim}.} \bibinfo{year}{2020}\natexlab{}.
\newblock \showarticletitle{FISR: Deep Joint Frame Interpolation and
  Super-Resolution with a Multi-Scale Temporal Loss}.
\newblock \bibinfo{journal}{\emph{Proceedings of the AAAI Conference on
  Artificial Intelligence}} (\bibinfo{date}{Apr.} \bibinfo{year}{2020}),
  \bibinfo{pages}{11278--11286}.
\newblock


\bibitem[Kong et~al\mbox{.}(2022)]%
        {Kong_2022_CVPR}
\bibfield{author}{\bibinfo{person}{Lingtong Kong}, \bibinfo{person}{Boyuan
  Jiang}, \bibinfo{person}{Donghao Luo}, \bibinfo{person}{Wenqing Chu},
  \bibinfo{person}{Xiaoming Huang}, \bibinfo{person}{Ying Tai},
  \bibinfo{person}{Chengjie Wang}, {and} \bibinfo{person}{Jie Yang}.}
  \bibinfo{year}{2022}\natexlab{}.
\newblock \showarticletitle{IFRNet: Intermediate Feature Refine Network for
  Efficient Frame Interpolation}. In \bibinfo{booktitle}{\emph{Proceedings of
  the IEEE/CVF Conference on Computer Vision and Pattern Recognition (CVPR)}}.
  \bibinfo{pages}{1969--1978}.
\newblock


\bibitem[Kong et~al\mbox{.}(2021)]%
        {Kong_2021_CVPR}
\bibfield{author}{\bibinfo{person}{Xiangtao Kong}, \bibinfo{person}{Hengyuan
  Zhao}, \bibinfo{person}{Yu Qiao}, {and} \bibinfo{person}{Chao Dong}.}
  \bibinfo{year}{2021}\natexlab{}.
\newblock \showarticletitle{ClassSR: A General Framework to Accelerate
  Super-Resolution Networks by Data Characteristic}. In
  \bibinfo{booktitle}{\emph{Proceedings of the IEEE/CVF Conference on Computer
  Vision and Pattern Recognition (CVPR)}}. \bibinfo{pages}{12016--12025}.
\newblock


\bibitem[Lee et~al\mbox{.}(2020)]%
        {Lee_2020_CVPR}
\bibfield{author}{\bibinfo{person}{Hyeongmin Lee}, \bibinfo{person}{Taeoh Kim},
  \bibinfo{person}{Tae-young Chung}, \bibinfo{person}{Daehyun Pak},
  \bibinfo{person}{Yuseok Ban}, {and} \bibinfo{person}{Sangyoun Lee}.}
  \bibinfo{year}{2020}\natexlab{}.
\newblock \showarticletitle{AdaCoF: Adaptive Collaboration of Flows for Video
  Frame Interpolation}. In \bibinfo{booktitle}{\emph{Proceedings of the
  IEEE/CVF Conference on Computer Vision and Pattern Recognition (CVPR)}}.
\newblock


\bibitem[Liu et~al\mbox{.}(2022)]%
        {Liu_2022_CVPR}
\bibfield{author}{\bibinfo{person}{Zhenhua Liu}, \bibinfo{person}{Yunhe Wang},
  \bibinfo{person}{Kai Han}, \bibinfo{person}{Siwei Ma}, {and}
  \bibinfo{person}{Wen Gao}.} \bibinfo{year}{2022}\natexlab{}.
\newblock \showarticletitle{Instance-Aware Dynamic Neural Network
  Quantization}. In \bibinfo{booktitle}{\emph{Proceedings of the IEEE/CVF
  Conference on Computer Vision and Pattern Recognition (CVPR)}}.
  \bibinfo{pages}{12434--12443}.
\newblock


\bibitem[Liu et~al\mbox{.}(2017)]%
        {Liu_2017_ICCV}
\bibfield{author}{\bibinfo{person}{Ziwei Liu}, \bibinfo{person}{Raymond~A.
  Yeh}, \bibinfo{person}{Xiaoou Tang}, \bibinfo{person}{Yiming Liu}, {and}
  \bibinfo{person}{Aseem Agarwala}.} \bibinfo{year}{2017}\natexlab{}.
\newblock \showarticletitle{Video Frame Synthesis Using Deep Voxel Flow}. In
  \bibinfo{booktitle}{\emph{Proceedings of the IEEE International Conference on
  Computer Vision (ICCV)}}.
\newblock


\bibitem[Loshchilov and Hutter(2019)]%
        {Loshchilov2019DecoupledWD}
\bibfield{author}{\bibinfo{person}{Ilya Loshchilov} {and}
  \bibinfo{person}{Frank Hutter}.} \bibinfo{year}{2019}\natexlab{}.
\newblock \showarticletitle{Decoupled Weight Decay Regularization}. In
  \bibinfo{booktitle}{\emph{ICLR}}.
\newblock


\bibitem[Lu et~al\mbox{.}(2022)]%
        {Lu_2022_CVPR}
\bibfield{author}{\bibinfo{person}{Liying Lu}, \bibinfo{person}{Ruizheng Wu},
  \bibinfo{person}{Huaijia Lin}, \bibinfo{person}{Jiangbo Lu}, {and}
  \bibinfo{person}{Jiaya Jia}.} \bibinfo{year}{2022}\natexlab{}.
\newblock \showarticletitle{Video Frame Interpolation With Transformer}. In
  \bibinfo{booktitle}{\emph{Proceedings of the IEEE/CVF Conference on Computer
  Vision and Pattern Recognition (CVPR)}}. \bibinfo{pages}{3532--3542}.
\newblock


\bibitem[Meister et~al\mbox{.}(2018)]%
        {Meister_Hur_Roth_2018}
\bibfield{author}{\bibinfo{person}{Simon Meister}, \bibinfo{person}{Junhwa
  Hur}, {and} \bibinfo{person}{Stefan Roth}.} \bibinfo{year}{2018}\natexlab{}.
\newblock \showarticletitle{UnFlow: Unsupervised Learning of Optical Flow With
  a Bidirectional Census Loss}.
\newblock \bibinfo{journal}{\emph{Proceedings of the AAAI Conference on
  Artificial Intelligence}} \bibinfo{volume}{32}, \bibinfo{number}{1}
  (\bibinfo{date}{Apr.} \bibinfo{year}{2018}).
\newblock


\bibitem[Mullapudi et~al\mbox{.}(2018)]%
        {Mullapudi_2018_CVPR}
\bibfield{author}{\bibinfo{person}{Ravi~Teja Mullapudi},
  \bibinfo{person}{William~R. Mark}, \bibinfo{person}{Noam Shazeer}, {and}
  \bibinfo{person}{Kayvon Fatahalian}.} \bibinfo{year}{2018}\natexlab{}.
\newblock \showarticletitle{HydraNets: Specialized Dynamic Architectures for
  Efficient Inference}. In \bibinfo{booktitle}{\emph{Proceedings of the IEEE
  Conference on Computer Vision and Pattern Recognition (CVPR)}}.
\newblock


\bibitem[Niklaus and Liu(2018)]%
        {Niklaus_2018_CVPR}
\bibfield{author}{\bibinfo{person}{Simon Niklaus} {and} \bibinfo{person}{Feng
  Liu}.} \bibinfo{year}{2018}\natexlab{}.
\newblock \showarticletitle{Context-Aware Synthesis for Video Frame
  Interpolation}. In \bibinfo{booktitle}{\emph{Proceedings of the IEEE
  Conference on Computer Vision and Pattern Recognition (CVPR)}}.
\newblock


\bibitem[Niklaus and Liu(2020)]%
        {Niklaus_2020_CVPR}
\bibfield{author}{\bibinfo{person}{Simon Niklaus} {and} \bibinfo{person}{Feng
  Liu}.} \bibinfo{year}{2020}\natexlab{}.
\newblock \showarticletitle{Softmax Splatting for Video Frame Interpolation}.
  In \bibinfo{booktitle}{\emph{Proceedings of the IEEE/CVF Conference on
  Computer Vision and Pattern Recognition (CVPR)}}.
\newblock


\bibitem[Niklaus et~al\mbox{.}(2017a)]%
        {Niklaus_2017_CVPR}
\bibfield{author}{\bibinfo{person}{Simon Niklaus}, \bibinfo{person}{Long Mai},
  {and} \bibinfo{person}{Feng Liu}.} \bibinfo{year}{2017}\natexlab{a}.
\newblock \showarticletitle{Video Frame Interpolation via Adaptive
  Convolution}. In \bibinfo{booktitle}{\emph{Proceedings of the IEEE Conference
  on Computer Vision and Pattern Recognition (CVPR)}}.
\newblock


\bibitem[Niklaus et~al\mbox{.}(2017b)]%
        {Niklaus_2017_ICCV}
\bibfield{author}{\bibinfo{person}{Simon Niklaus}, \bibinfo{person}{Long Mai},
  {and} \bibinfo{person}{Feng Liu}.} \bibinfo{year}{2017}\natexlab{b}.
\newblock \showarticletitle{Video Frame Interpolation via Adaptive Separable
  Convolution}. In \bibinfo{booktitle}{\emph{Proceedings of the IEEE
  International Conference on Computer Vision (ICCV)}}.
\newblock


\bibitem[Ning et~al\mbox{.}(2021)]%
        {NEURIPS2021_88a19961}
\bibfield{author}{\bibinfo{person}{Qian Ning}, \bibinfo{person}{Weisheng Dong},
  \bibinfo{person}{Xin Li}, \bibinfo{person}{Jinjian Wu}, {and}
  \bibinfo{person}{GUANGMING Shi}.} \bibinfo{year}{2021}\natexlab{}.
\newblock \showarticletitle{Uncertainty-Driven Loss for Single Image
  Super-Resolution}. In \bibinfo{booktitle}{\emph{Advances in Neural
  Information Processing Systems}}, Vol.~\bibinfo{volume}{34}.
  \bibinfo{publisher}{Curran Associates, Inc.}, \bibinfo{pages}{16398--16409}.
\newblock


\bibitem[Parger et~al\mbox{.}(2022)]%
        {Parger_2022_CVPR}
\bibfield{author}{\bibinfo{person}{Mathias Parger}, \bibinfo{person}{Chengcheng
  Tang}, \bibinfo{person}{Christopher~D. Twigg}, \bibinfo{person}{Cem Keskin},
  \bibinfo{person}{Robert Wang}, {and} \bibinfo{person}{Markus Steinberger}.}
  \bibinfo{year}{2022}\natexlab{}.
\newblock \showarticletitle{DeltaCNN: End-to-End CNN Inference of Sparse Frame
  Differences in Videos}. In \bibinfo{booktitle}{\emph{Proceedings of the
  IEEE/CVF Conference on Computer Vision and Pattern Recognition (CVPR)}}.
  \bibinfo{pages}{12497--12506}.
\newblock


\bibitem[Park et~al\mbox{.}(2020)]%
        {10.1007/978-3-030-58568-6_7}
\bibfield{author}{\bibinfo{person}{Junheum Park}, \bibinfo{person}{Keunsoo Ko},
  \bibinfo{person}{Chul Lee}, {and} \bibinfo{person}{Chang-Su Kim}.}
  \bibinfo{year}{2020}\natexlab{}.
\newblock \showarticletitle{BMBC: Bilateral Motion Estimation with Bilateral
  Cost Volume for Video Interpolation}. In \bibinfo{booktitle}{\emph{Computer
  Vision -- ECCV 2020}}, \bibfield{editor}{\bibinfo{person}{Andrea Vedaldi},
  \bibinfo{person}{Horst Bischof}, \bibinfo{person}{Thomas Brox}, {and}
  \bibinfo{person}{Jan-Michael Frahm}} (Eds.). \bibinfo{pages}{109--125}.
\newblock


\bibitem[Park et~al\mbox{.}(2021)]%
        {Park_2021_ICCV}
\bibfield{author}{\bibinfo{person}{Junheum Park}, \bibinfo{person}{Chul Lee},
  {and} \bibinfo{person}{Chang-Su Kim}.} \bibinfo{year}{2021}\natexlab{}.
\newblock \showarticletitle{Asymmetric Bilateral Motion Estimation for Video
  Frame Interpolation}. In \bibinfo{booktitle}{\emph{Proceedings of the
  IEEE/CVF International Conference on Computer Vision (ICCV)}}.
  \bibinfo{pages}{14539--14548}.
\newblock


\bibitem[Peleg et~al\mbox{.}(2019)]%
        {Peleg_2019_CVPR}
\bibfield{author}{\bibinfo{person}{Tomer Peleg}, \bibinfo{person}{Pablo
  Szekely}, \bibinfo{person}{Doron Sabo}, {and} \bibinfo{person}{Omry Sendik}.}
  \bibinfo{year}{2019}\natexlab{}.
\newblock \showarticletitle{IM-Net for High Resolution Video Frame
  Interpolation}. In \bibinfo{booktitle}{\emph{Proceedings of the IEEE/CVF
  Conference on Computer Vision and Pattern Recognition (CVPR)}}.
\newblock


\bibitem[Reda et~al\mbox{.}(2018)]%
        {Reda_2018_ECCV}
\bibfield{author}{\bibinfo{person}{Fitsum~A. Reda}, \bibinfo{person}{Guilin
  Liu}, \bibinfo{person}{Kevin~J. Shih}, \bibinfo{person}{Robert Kirby},
  \bibinfo{person}{Jon Barker}, \bibinfo{person}{David Tarjan},
  \bibinfo{person}{Andrew Tao}, {and} \bibinfo{person}{Bryan Catanzaro}.}
  \bibinfo{year}{2018}\natexlab{}.
\newblock \showarticletitle{SDC-Net: Video prediction using spatially-displaced
  convolution}. In \bibinfo{booktitle}{\emph{Proceedings of the European
  Conference on Computer Vision (ECCV)}}.
\newblock


\bibitem[Shi et~al\mbox{.}(2022)]%
        {Shi_2022_CVPR}
\bibfield{author}{\bibinfo{person}{Zhihao Shi}, \bibinfo{person}{Xiangyu Xu},
  \bibinfo{person}{Xiaohong Liu}, \bibinfo{person}{Jun Chen}, {and}
  \bibinfo{person}{Ming-Hsuan Yang}.} \bibinfo{year}{2022}\natexlab{}.
\newblock \showarticletitle{Video Frame Interpolation Transformer}. In
  \bibinfo{booktitle}{\emph{Proceedings of the IEEE/CVF Conference on Computer
  Vision and Pattern Recognition (CVPR)}}. \bibinfo{pages}{17482--17491}.
\newblock


\bibitem[Soomro et~al\mbox{.}(2012)]%
        {ucf101}
\bibfield{author}{\bibinfo{person}{Khurram Soomro}, \bibinfo{person}{Amir
  Zamir}, {and} \bibinfo{person}{Mubarak Shah}.}
  \bibinfo{year}{2012}\natexlab{}.
\newblock \showarticletitle{UCF101: A Dataset of 101 Human Actions Classes From
  Videos in The Wild}.
\newblock \bibinfo{journal}{\emph{CoRR}} (\bibinfo{date}{12}
  \bibinfo{year}{2012}).
\newblock


\bibitem[Wang et~al\mbox{.}(2021)]%
        {Wang_2021_CVPR}
\bibfield{author}{\bibinfo{person}{Longguang Wang}, \bibinfo{person}{Xiaoyu
  Dong}, \bibinfo{person}{Yingqian Wang}, \bibinfo{person}{Xinyi Ying},
  \bibinfo{person}{Zaiping Lin}, \bibinfo{person}{Wei An}, {and}
  \bibinfo{person}{Yulan Guo}.} \bibinfo{year}{2021}\natexlab{}.
\newblock \showarticletitle{Exploring Sparsity in Image Super-Resolution for
  Efficient Inference}. In \bibinfo{booktitle}{\emph{Proceedings of the
  IEEE/CVF Conference on Computer Vision and Pattern Recognition (CVPR)}}.
  \bibinfo{pages}{4917--4926}.
\newblock


\bibitem[Wu et~al\mbox{.}(2018b)]%
        {Wu_2018_ECCV}
\bibfield{author}{\bibinfo{person}{Chao-Yuan Wu}, \bibinfo{person}{Nayan
  Singhal}, {and} \bibinfo{person}{Philipp Krahenbuhl}.}
  \bibinfo{year}{2018}\natexlab{b}.
\newblock \showarticletitle{Video Compression through Image Interpolation}. In
  \bibinfo{booktitle}{\emph{Proceedings of the European Conference on Computer
  Vision (ECCV)}}.
\newblock


\bibitem[Wu et~al\mbox{.}(2018a)]%
        {Wu_2018_CVPR}
\bibfield{author}{\bibinfo{person}{Zuxuan Wu}, \bibinfo{person}{Tushar
  Nagarajan}, \bibinfo{person}{Abhishek Kumar}, \bibinfo{person}{Steven
  Rennie}, \bibinfo{person}{Larry~S. Davis}, \bibinfo{person}{Kristen Grauman},
  {and} \bibinfo{person}{Rogerio Feris}.} \bibinfo{year}{2018}\natexlab{a}.
\newblock \showarticletitle{BlockDrop: Dynamic Inference Paths in Residual
  Networks}. In \bibinfo{booktitle}{\emph{Proceedings of the IEEE Conference on
  Computer Vision and Pattern Recognition (CVPR)}}.
\newblock


\bibitem[Xie et~al\mbox{.}(2020)]%
        {10.1007/978-3-030-58452-8_31}
\bibfield{author}{\bibinfo{person}{Zhenda Xie}, \bibinfo{person}{Zheng Zhang},
  \bibinfo{person}{Xizhou Zhu}, \bibinfo{person}{Gao Huang}, {and}
  \bibinfo{person}{Stephen Lin}.} \bibinfo{year}{2020}\natexlab{}.
\newblock \showarticletitle{Spatially Adaptive Inference with Stochastic
  Feature Sampling and Interpolation}. In \bibinfo{booktitle}{\emph{Computer
  Vision -- ECCV 2020}}. \bibinfo{address}{Cham}, \bibinfo{pages}{531--548}.
\newblock


\bibitem[Xing et~al\mbox{.}(2020)]%
        {10.1007/978-3-030-58517-4_17}
\bibfield{author}{\bibinfo{person}{Qunliang Xing}, \bibinfo{person}{Mai Xu},
  \bibinfo{person}{Tianyi Li}, {and} \bibinfo{person}{Zhenyu Guan}.}
  \bibinfo{year}{2020}\natexlab{}.
\newblock \showarticletitle{Early Exit or Not: Resource-Efficient Blind Quality
  Enhancement for Compressed Images}. In \bibinfo{booktitle}{\emph{Computer
  Vision -- ECCV 2020}}. \bibinfo{pages}{275--292}.
\newblock


\bibitem[Xu et~al\mbox{.}(2019)]%
        {NEURIPS2019_d045c59a}
\bibfield{author}{\bibinfo{person}{Xiangyu Xu}, \bibinfo{person}{Li Siyao},
  \bibinfo{person}{Wenxiu Sun}, \bibinfo{person}{Qian Yin}, {and}
  \bibinfo{person}{Ming-Hsuan Yang}.} \bibinfo{year}{2019}\natexlab{}.
\newblock \showarticletitle{Quadratic Video Interpolation}. In
  \bibinfo{booktitle}{\emph{Advances in Neural Information Processing
  Systems}}, Vol.~\bibinfo{volume}{32}. \bibinfo{publisher}{Curran Associates,
  Inc.}
\newblock


\bibitem[Xu et~al\mbox{.}(2022)]%
        {Xu_Wang_Li_Lu_2022}
\bibfield{author}{\bibinfo{person}{Xiaohao Xu}, \bibinfo{person}{Jinglu Wang},
  \bibinfo{person}{Xiao Li}, {and} \bibinfo{person}{Yan Lu}.}
  \bibinfo{year}{2022}\natexlab{}.
\newblock \showarticletitle{Reliable Propagation-Correction Modulation for
  Video Object Segmentation}.
\newblock \bibinfo{journal}{\emph{Proceedings of the AAAI Conference on
  Artificial Intelligence}} \bibinfo{volume}{36}, \bibinfo{number}{3}
  (\bibinfo{date}{Jun.} \bibinfo{year}{2022}), \bibinfo{pages}{2946--2954}.
\newblock
\urldef\tempurl%
\url{https://doi.org/10.1609/aaai.v36i3.20200}
\showDOI{\tempurl}


\bibitem[Yang et~al\mbox{.}(2022)]%
        {Yang_2022_CVPR}
\bibfield{author}{\bibinfo{person}{Chenhongyi Yang}, \bibinfo{person}{Zehao
  Huang}, {and} \bibinfo{person}{Naiyan Wang}.}
  \bibinfo{year}{2022}\natexlab{}.
\newblock \showarticletitle{QueryDet: Cascaded Sparse Query for Accelerating
  High-Resolution Small Object Detection}. In
  \bibinfo{booktitle}{\emph{Proceedings of the IEEE/CVF Conference on Computer
  Vision and Pattern Recognition (CVPR)}}. \bibinfo{pages}{13668--13677}.
\newblock


\bibitem[Zhou et~al\mbox{.}(2016)]%
        {10.1007/978-3-319-46493-0_18}
\bibfield{author}{\bibinfo{person}{Tinghui Zhou}, \bibinfo{person}{Shubham
  Tulsiani}, \bibinfo{person}{Weilun Sun}, \bibinfo{person}{Jitendra Malik},
  {and} \bibinfo{person}{Alexei~A. Efros}.} \bibinfo{year}{2016}\natexlab{}.
\newblock \showarticletitle{View Synthesis by Appearance Flow}. In
  \bibinfo{booktitle}{\emph{Computer Vision -- ECCV 2016}}.
  \bibinfo{pages}{286--301}.
\newblock


\end{thebibliography}

\appendix

\section{Implementation Details}
\label{appendix:implementation}
Our UGSP consists of two training phases. In the first phase, we train the UEN network to predict the uncertainty, and in the second phase, we use the uncertainty to guide the training of the VFI network.
We choose the AdamW \cite{Loshchilov2019DecoupledWD} optimizer for 100 and 300 epochs to train the UEN and VFI networks.
UGSP is implemented with Pytorch and trained on four NVIDIA GeForce RTX 3090 GPUs with a batch size of 32 and a patch size of 224$\times$224. 
The learning rate decays from 1e-4 to 1e-5 through cosine annealing.
We set $\alpha_1\%$, $\alpha_2\%$, and $\alpha_3\%$ to 20\%, 40\%, and 80\%, respectively, to generate pruning mask label in Equation~\ref{equ:label} when the target sparsity $S_t$ is 35\% in the sparse loss (Equation~\ref{equ:s}).
The weights $\lambda_{s}$, $\lambda_{u g m}$, and $\lambda_{{s c}}$ in the overall loss (Equation~\ref{equ:overall}) are set to 0.01.
We used the sparse convolution implemented by Wang \emph{et al.} \cite{Wang_2021_CVPR}.

\section{Limitation}
Currently, the decreases in FLOPs do not bring faster GPU inference times, as shown in Table~\ref{tab:limitation}.
As discussed in Wang \emph{et al.} \cite{Wang_2021_CVPR}, because of the irregular and fragmented memory patterns, sparse convolution in UGSP cannot fully exploit GPUs without specific optimization.
Therefore, we intend to explore sparse convolution friendly optimizations in the future.

\begin{table}[h]
    \small
    \begin{tabular}{lccc}

    \hline
    & Baseline & UGSP-large
    & UGSP \\
    \hline
    FLOPs & 31.7G & 21.0G & 15.6G\\
    CPU &  1.16s & 0.59s & 0.45s \\
    GPU &  9.4ms & 14.2ms & 14.0ms \\
    \hline
    \end{tabular}
    \caption{ Comparison of FLOPs and GPU time on an NVIDIA GeForce RTX 3090 GPU on the Vimeo90K dataset. }
    \label{tab:limitation}
\end{table}

\section{Algorithm of UGSP Framework}
\label{appendix:algorithm}
We summarize our two phases of UGSP training in Algorithm~\ref{alg:ugsp}. 
The uncertainty estimation network (UEN) in the first phase is trained using the Vimeo90K training dataset. UEN can predict the uncertainty (variance) field $U_k$ for the unknown intermediate frame. The uncertainty map is then used to generate pruning masks $P_k^u$ that serve as a guide for the second phase's spatial pruning mask estimation using $\mathcal{L}_{ugm}$.
Therefore, we save the $P_k^u$ for each Vimeo90K training dataset sample at the end of first phase. 

In the second phase, we train the VFI network using the Vimeo90K training dataset and $P_k^u$.
$\mathcal{L}_s$ controls the degree of sparsity in our UGSP, and $\mathcal{L}_{sc}$ utilizes the feature of the auxiliary non-pruning branch to enhance the performance of our UGSP.
Finally, we only obtain the VFI nework $E_{vfi}$, and the UEN $E_{uen}$ is not required during the inference phase.

\begin{algorithm}[h]
	\caption{Two training phases of UGSP.}
	\label{alg:ugsp}
	\KwIn{Initialize network parameters $\theta_{uen}$ and $\theta_{vfi}$ of UEN $E_{uen}$ and VFI network $E_{vfi}$, respectively. Set $\lambda_{s}$=$\lambda_{ugm}$=$\lambda_{sc}$=0.01.}
	\BlankLine

        \# First phase:
        
	\While{\textnormal{$\theta_{uen}$ has not converged}}{
    	 Sample $\{I_0, I_1\}$ a batch from the Vimeo90K training dataset; \\
         $\{I_t^k, U_k\} \leftarrow E_{uen}(I_0, I_1), k\in(0,1,2)$; \\
         Calculate $\mathcal{L}_{s u}(I_t^k, U_k), k\in(0,1,2)$;
        $\theta_{esu} \leftarrow \theta_{esu}- \nabla_{\theta_{esu}} L_{s u}$;
	}
        \ForEach{$\{I_0, I_1\}$ in Vimeo90K training dataset}{
		$\{U_k\} \leftarrow E_{uen}(I_0, I_1), k\in(0,1,2)$; \\
            $\{P_k^u\} \leftarrow \{U_k\}, k\in(0,1,2)$ using Equation (4); \\
            Save $\{P_k^u\}, k\in(0,1,2)$;
		}

	\BlankLine
        \# Second phase:
        
	\While{\textnormal{$\theta_{vfi}$ has not converged}}{
    	 Sample $\{I_0, I_1, \{P_k^u\}, k\in(0,1,2) \}$ a batch from the Vimeo90K training dataset; \\
         $\{I_t, I_t^{sc}\} \leftarrow E_{vfi}(I_0, I_1, \{P_k^u\}, k\in(0,1,2))$; \\
         Calculate $\mathcal{L}_{r e c}(I_t, I_t^{gt})$, $\mathcal{L}_{sc}(I_t, I_t^{sc}, I_t^{gt})$; \\
         Calculate $\mathcal{L}_{s}(P_k), \mathcal{L}_{ugm}(P_k, P_k^u), k\in(0,1,2)$; \\
        $\theta_{vfi} \leftarrow \theta_{vfi}- \nabla_{\theta_{vfi}} \mathcal{L}_{rec}$;\\
        $\theta_{vfi} \leftarrow \theta_{vfi}- \nabla_{\theta_{vfi}} \lambda_{s}\mathcal{L}_{s}$;\\
        $\theta_{vfi} \leftarrow \theta_{vfi}- \nabla_{\theta_{vfi}} \lambda_{ugm}\mathcal{L}_{ugm}$;\\
        $\theta_{vfi} \leftarrow \theta_{vfi}- \nabla_{\theta_{vfi}} \lambda_{sc}\mathcal{L}_{sc}$;\\
	}
    \BlankLine
    \KwOut{VFI network $E_{vfi}$.}  

\end{algorithm}

\begin{figure}[h]
\centering
\includegraphics[width=\linewidth]{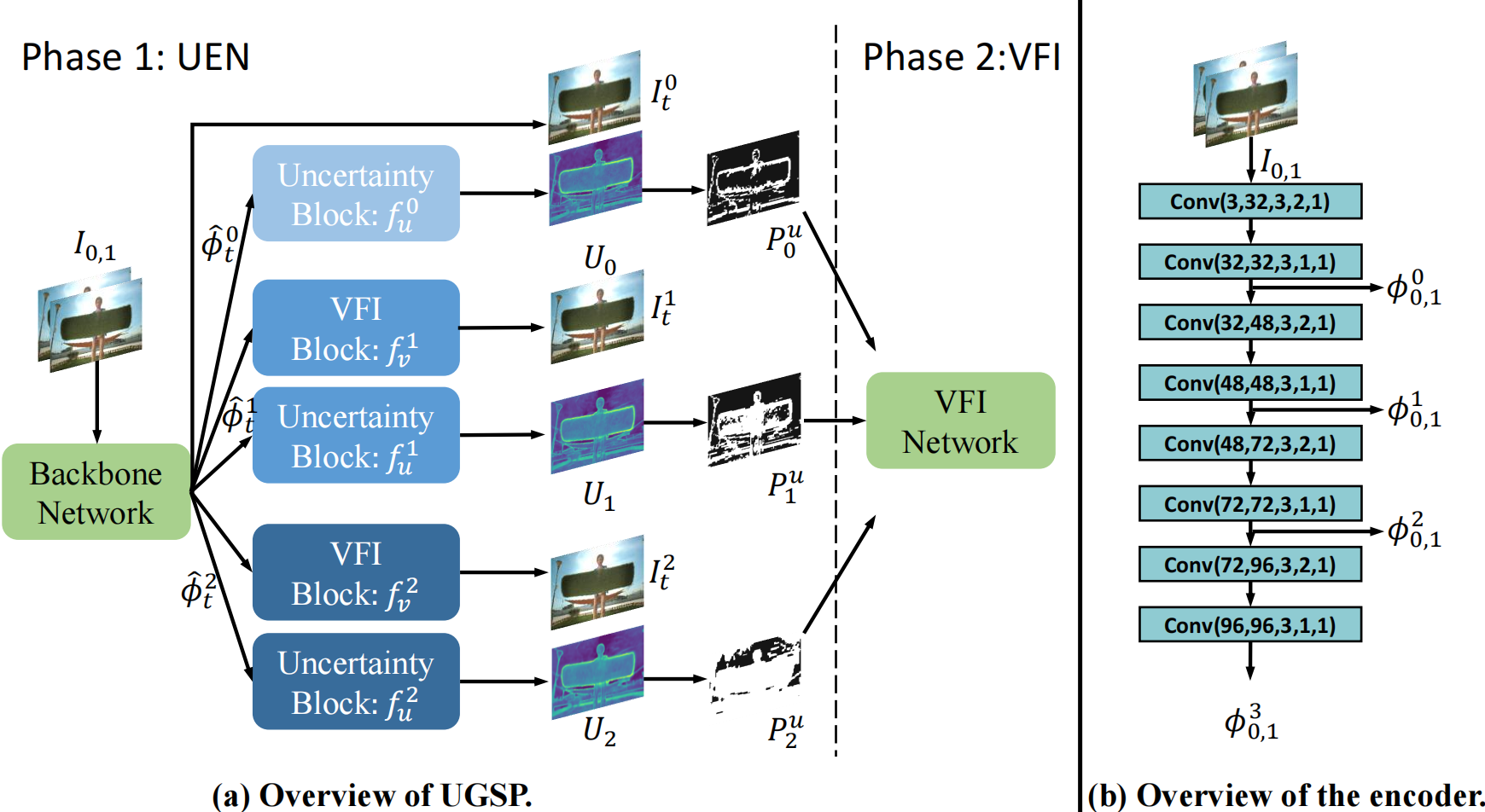}
\caption{Overview of our UGSP and encoder. The PReLu activation following each convolution has been omitted for clarity.}
\label{s_overview_encoder}
\end{figure}



\begin{figure*}[h]
\centering
\includegraphics[width=\linewidth]{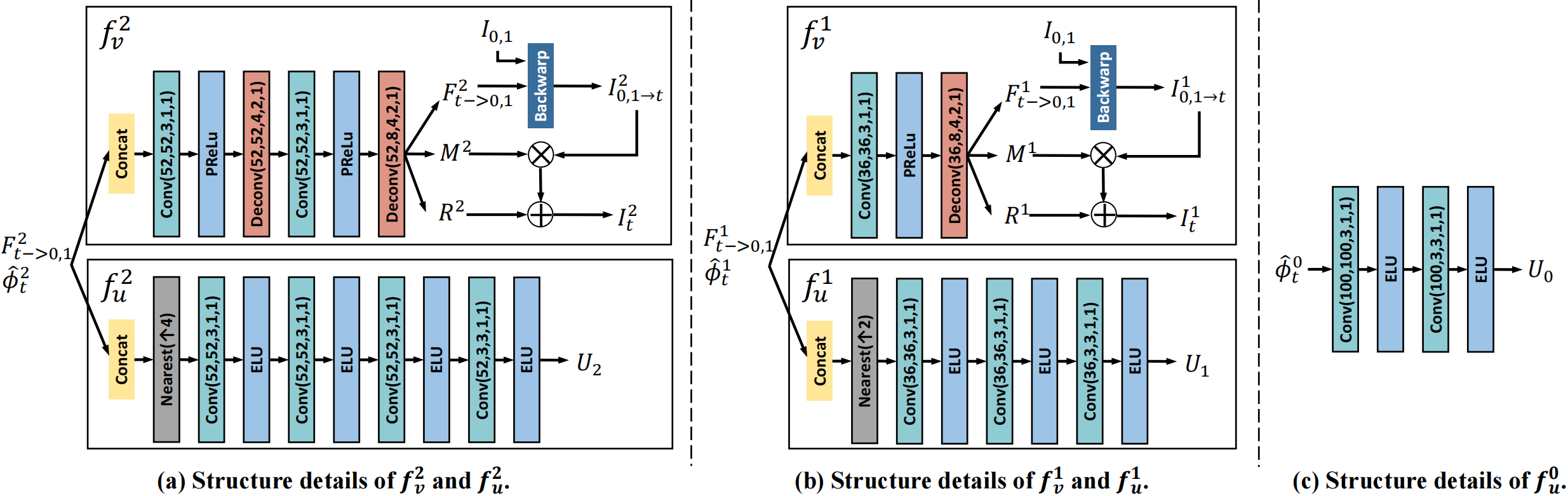}
\caption{Overview of the Uncertainty blocks ($f_u^0$, $f_u^1$ and $f_u^2$) and VFI blocks ($f_v^1$ and $f_v^2$) in UEN. }
\label{s_uen}
\end{figure*}

\begin{figure*}[h]
\centering
\includegraphics[width=\linewidth]{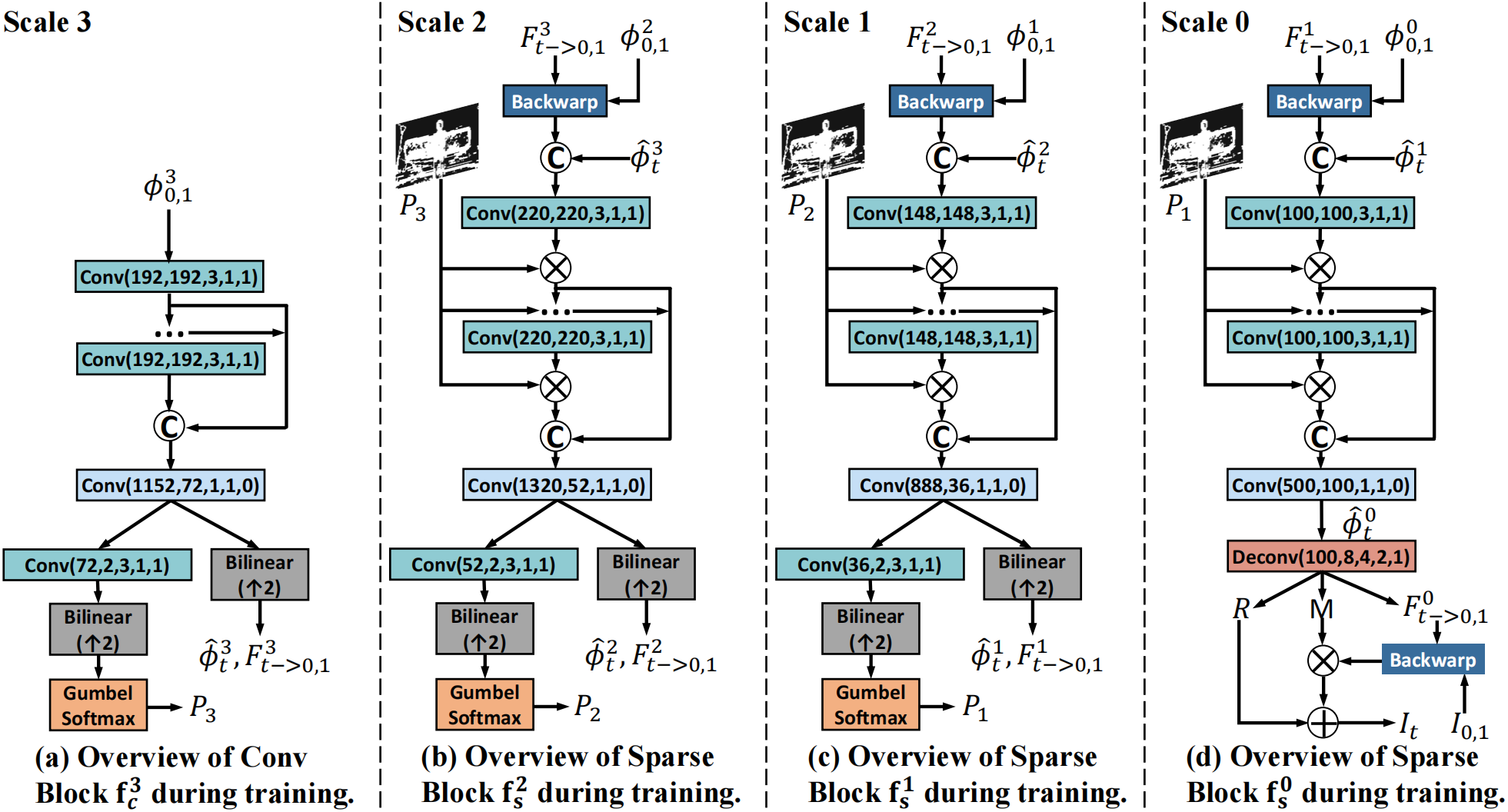}
\caption{Overview of the VFI network in each scale during training. The PReLu activation following each 3$\times$3 convolution has been omitted for clarity.}
\label{s_vfi}
\end{figure*}

\section{Network Architecture}
\label{appendix:network}
This section illustrates all VFI and UEN structure details, and the overview of our UGSP is shown in Figure~\ref{s_overview_encoder}(a).
In Figure~\ref{s_uen}, we display the structure details of Uncertainty blocks ($f_u^0$, $f_u^1$ and $f_u^2$) and VFI blocks ($f_v^1$ and $f_v^2$).
Uncertainty blocks ($f_u^0$, $f_u^1$ and $f_u^2$) first use one nearest up-sampling operation to maintain same resolution as the input frame. Then, they respectively estimate uncertainty ($U_0$, $U_1$, and $U_2$) using 2, 3, and 4 blocks of one $3\times3$ convolution and one ELU activation layer. VFI blocks ($f_v^1$ and $f_v^2$) respectively contain 1 and 2 blocks of one $3\times3$ convolution, one PReLu activation layer, and one deconvolution to estimate the middle frame ($I_t^1$ and $I_t^2$). Therefore, the resolution of $U_k$ and $I_t^k$ is the same as the input frame $I_{0,1}$.

Since the backbone network structure in UEN is identical to that in the VFI network, but removes a branch for estimating the pruning mask and adding a brach for estimationg uncertainty. Therefore, we will only describe the structure of the VFI network.
As shown in Figure~\ref{s_overview_encoder}(b), VFI first down-samples two input frames $I_0$ and $I_1$, four times using a block of two 3×3 convolutions with strides 2 and 1 to obtain four level features with 32, 48, 72, and 96 channels $\phi_{0,1}^k, k \in\{0,1,2,3\}$ in the encoder. 

In Figure~\ref{s_vfi}, we illustrate the process in each scale of VFI network. 
As shown in Figure~\ref{s_vfi}(a), we concatenate the output of six consecutive $3\times3$ convolutions and then use one $1\times1$ convolution to extract features.
$F_{t \rightarrow 0,1}^3$ and $\hat{\phi}_t^3$ are obtained via one bilinear up-sampling, and $P_3$ is obtained via one $3\times3$ convolution, one bilinear up-sampling, and a Gumbel softmax.
The Gumbel softmax trick is applied to obtain a softened spatial mask for $P^k, k\in{1,2,3}$:
\begin{equation}
P^k=\frac{\exp \left(\left(\hat{P}^{k}[1]+G^k[1]\right) / \tau\right)}{\sum_{i=1}^2 \exp \left(\left(\hat{P}^{k}[i]+G^k[i]\right) / \tau\right)},
\end{equation}
where ${P}^{k}$ is the output of the bilinear up-sampling. $G^k$ is a Gumbel noise tensor where elements all have a Gumbel(0, 1) distribution, and $\tau$ is a temperature parameter. In practice, we generate binary pruning masks by starting at a high temperature and annealing to a low one.
$P_{3}$ is used to skip the redundant computations in the $3\times3$ and $1\times1$ convolutions of scale 2. 

As shown in Figure~\ref{s_vfi}(b-d), we achieve sparse convolution by multiplying each convolution operation's output feature with the pruning mask predicted from the previous scale during training.
We first concatenate the feature $\hat{\phi}_{t}^k$ and the aligned feature produced by back warping using flow field $F_{t \rightarrow 0,1}^k$ and feature $\phi_{0,1}^{k-1}$.
Then we input it into consecutive $3\times3$ convolutions and one $1\times1$ convolution to refine the feature.
In scales 1 and 2, $(F_{t \rightarrow 0,1}^{k-1}, \hat{\phi}_t^{k-1}, P_{k-1})$ are achieved in the same manner as in scale 3.
In scale 0, we utilize one deconvolution to achieve $F_{t \rightarrow 0,1}^0, M, R$, which are then blended into the estimated intermediate frame $I_t$.

\section{Details of UGSP-\textit{distill} and UGSP-\textit{refine}}
\label{appendix:ugsp}

This section describes the structure of UGSP-\textit{distill} and UGSP-\textit{refine}.

\noindent
\textbf{UGSP-\textit{distill}.} We implement the flow distillation and geometry consistency loss of IFRnet \cite{Kong_2022_CVPR} into UGSP as UGSP-\textit{distill}.
For the flow distillation loss, the pre-trained liteflownet \footnote{http://content.sniklaus.com/github/pytorch-liteflownet/network-default.pytorch} is used to predict the pseudo flow label $F_{t\rightarrow0}^p$,  $F_{t\rightarrow1}^p$.
Then,  we can obtain robustness masks
$P_l(l \in {0,1})$ by the following formulation:
\begin{equation}
P_l=\exp \left(-\beta\left|F_{t \rightarrow l}^{0}-F_{t \rightarrow l}^p\right|_{e p e}\right)
\end{equation}
where $\beta$ is set to 0.3 as IFRnet, and per-pixel end-point error is calculated between our estimated flow $F_{t \rightarrow l}^{0}$ in level 0 and the pseudo label $F_{t\rightarrow0}^p$.
Finally, we use the flow distillation loss, which can be expressed as:
\begin{equation}
\mathcal{L}_d=\sum_{k=1}^3 \sum_{l=0}^1 \left(\left(\left(F_{t \rightarrow l}^k\right) \uparrow_{2^k}-F_{t \rightarrow l}^{p}\right)^2 + 10^{-(10p-1)/3}\right)^{p/2},
\end{equation}
where $p$ denotes the value of any position in the mask $P_0$ and $P_1$. By task-oriented adjusting the Charbonnier loss $\rho(x)=\left(x^2+\epsilon^2\right)^\alpha$ using $p$, we can prevent the model from learning the pseudo label with noise.
$\uparrow_{2^k}$ denotes we bilinearly up-sample estimated flow to make the resolution consistent with the pseudo label.

    Geometry consistency loss can be expressed as:
    \begin{equation}
    \mathcal{L}_g=\sum_{k=1}^3 \mathcal{L}_{c e n}\left(\hat{\phi}_t^k, \phi_t^k\right),
    \end{equation}
where $\hat{\phi}_t^k$ denotes the output of the encoder when the ground truth intermediate frame is input.
$\mathcal{L}_{g}$ uses low-level structure information in $\hat{\phi}_t^k$ to regularize the reconstructed intermediate feature $\phi_t^k$.

\begin{figure}[h]
\centering
\includegraphics[width=\linewidth]{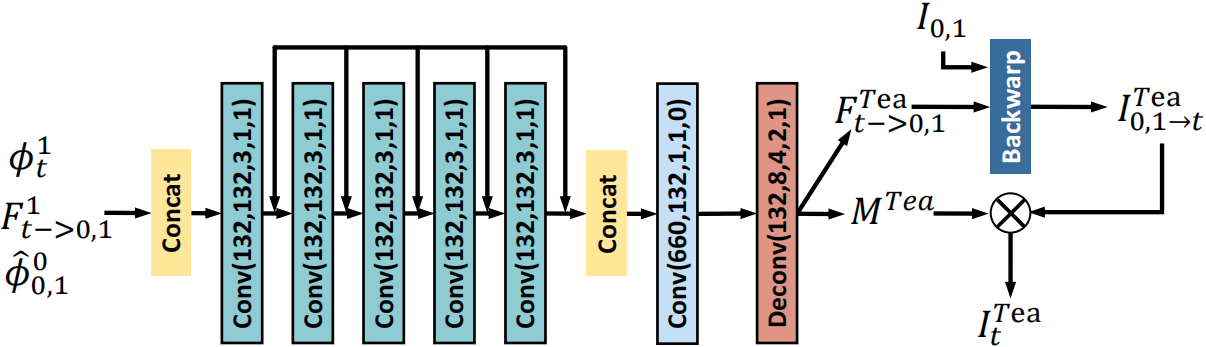}
\caption{Overview of the teacher model in UGSP-\textit{refine}. }
\label{s_rife}
\end{figure}

\noindent
\textbf{UGSP-\textit{refine}.}
We incorporate RIFE's \cite{huang2022rife} refinement network and priveleged distillation strategy into our framework as UGSP-\textit{refine}.
Specifically, we use the same refinement network in RIFE, but the channel is scaled by 0.625.
RIFE uses a teacher model to estimate optical flow by inputting the intermediate features of input frames and the ground truth intermediate frame $I_t^{gt}$.
Therefore, we also implement a teacher network, as shown in Figure~\ref{s_rife}. It is similar to the scale 0 procedure in the VFI network in Figure~\ref{s_vfi}(d) but without a pruning mask $P_1$.
Another difference is that we do not use the residual $R$ since the input contains the ground truth feature $\phi_t^{1}$, which would make the flow estimation inefficient if the residual is used.
$\phi_t^{1}$ is achieved by inputting the ground truth $I_t^{gt}$ into the encoder as shown in Figure~\ref{s_overview_encoder}(b).
Then, the distillation loss is defined as:
    \begin{equation}
    \mathcal{L}_{dis}=\mathcal{L}_{r e c} \left(I_t^{Tea}, I_t^{gt}  \right) + \sum_{k=0}^2 \left\|(F_{t\rightarrow{0,1}}^k)\uparrow_{2^k}- F_{t\rightarrow{0,1}}^{Tea}\right\|.
    \end{equation}
The teacher model obtains $I_t^{Tea}$ and $F_{t\rightarrow{0,1}}^{Tea}$ using $\phi_t^{1}$ as additional input.
$\uparrow_{2^k}$ denotes we bilinearly up-sample estimated flow to make resolution consistent with the teacher flow $F_{t\rightarrow{0,1}}^{Tea}$.
Like RIFE, the teacher block will be discarded after the training phase, so there would be no additional cost for inference.
$\mathcal{L}_{dis}$ not only makes more stable training but also enhances our model's estimation ability.

\begin{table}[h]
    \centering
    \resizebox{\linewidth}{!}
{
    \begin{tabular}{lcccccc}
    \toprule
   \multirow{2}{*}{ Model} &  \multicolumn{2}{c}{Vimeo90K} & \multicolumn{2}{c}{UCF101} &  \multicolumn{2}{c}{Middlebury} \\
     \cmidrule(r){2-3} \cmidrule(r){4-5} \cmidrule(r){6-7}  & FLOPs(G) & PSNR  & FLOPs(G) & PSNR &  FLOPs(G) & PSNR \\
    \hline
    10\%,60\%,80\% & 15.7 & 35.60 & 8.9 & 35.30 & 39.4 & 36.94 \\
    20\%,40\%,80\% & 15.6 & 35.62 & 8.6 & 35.31 & 39.3 & 36.96 \\
    30\%,30\%,60\% & 15.6 & 35.62 & 8.5 & 35.29 & 39.4 & 36.95 \\
    \hline
    \end{tabular}
}
    \caption{ 
  Ablation study of uncertainty map threshold.
}
    \label{tab:threshold}
\end{table}

\section{Ablation of Uncertainty Map Threshold}
\label{appendix:thresholding}

As described in Section~\ref{sec:ugm}, we assign the threshold $T_k$
 to the $\alpha_k$ smallest value of $U_{k-1}$. We set the threshold based on the convolution’s FLOPs in the pruning scale, and the FLOPs ratio of convolution on the scale 0, 1, 2 is close to 1: 2: 4, so we set $\alpha_1$, $\alpha_2$, $\alpha_3$ to 1: 2: 4. For example, $\alpha_1$, $\alpha_2$, $\alpha_3$ are assigned to 20\%, 40\%, 80\% when the target sparsity is 35\%. Here, we do the sensitivity ablation study for two models. The $\alpha_1$, $\alpha_2$, $\alpha_3$ for the first model are 10\%, 60\%, 80\%, and the 
$\alpha_1$, $\alpha_2$, $\alpha_3$ for the second model are 30\%, 30\%, 60\%. The results are shown in Table~\ref{tab:threshold}. 
We can observe that (20\%, 40\%, 80\%) model is better than (10\%, 60\%, 80\%) model. This might be because larger $\alpha_1$ forces more areas to use all computing resources, leading to more challenging regions to obtain all computing resources. In addition, (20\%, 40\%, 80\%) model obtains similar results to (30\%, 30\%, 60\%) model. The percentage of challenge regions might be less than 30\%, so increasing the number of regions utilizing all computing resources cannot significantly improve performance. We believe there are better ways to choose $\alpha_k$, and we will study it in the future. 

\begin{figure*}[h]
\centering
\includegraphics[width=\linewidth]{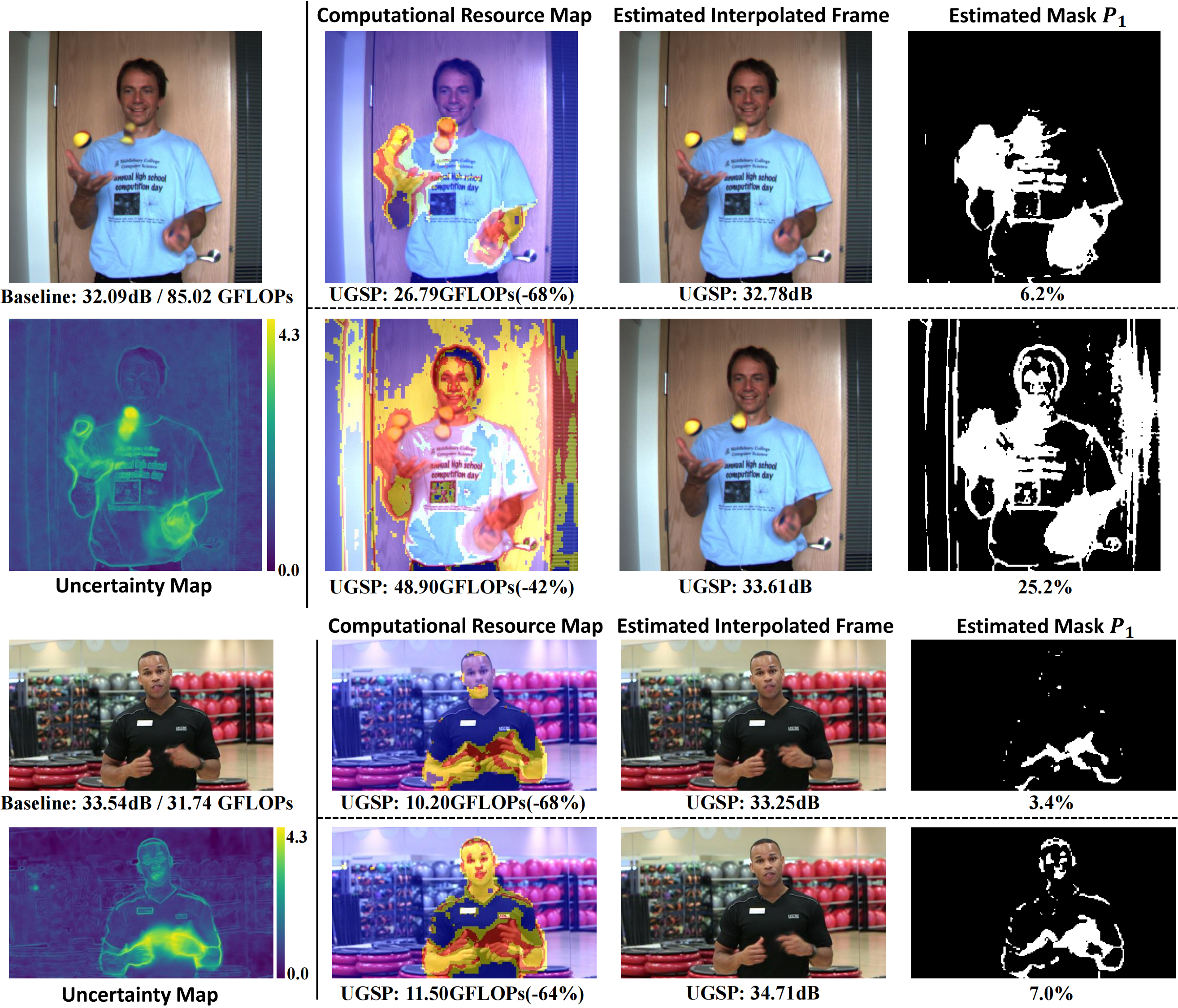}
\caption{Visual results of our computational resource map, instance overlaid, uncertainty map, and estimated mask $P_1$. The first example comes from the Middlebury \cite{Baker2011} Other dataset, and the second example comes from the Vimeo90K \cite{vimeo} testing datasets.}
\label{s_intro}
\end{figure*}

\section{More Visual Results}
\label{visual}
Figure~\ref{s_intro} illustrates the computational resource map for two examples, with the red, yellow, and blue regions requiring high, medium, and low computational costs, respectively.
It is observed that red regions nearly contain the region of large or complex movement, which is essential for a pleasing visual experience but challenging for VFI tasks.
Moreover, when we increase the computational resources, the PSNR increases, and complex and large motion is estimated more accurately .
Furthermore, our pruning model has a higher PSNR, and there are two reasons for this. First, using our estimated mask during training, our UGSP prioritizes the reconstructed quality of challenging areas.
Second, the estimation of simple and small motions in the baseline may harm large and complex motion estimation.

\end{document}